\documentclass[letterpaper]{article} % DO NOT CHANGE THIS
\usepackage{aaai25}  % DO NOT CHANGE THIS
\usepackage{times}  % DO NOT CHANGE THIS
\usepackage{helvet}  % DO NOT CHANGE THIS
\usepackage{courier}  % DO NOT CHANGE THIS
\usepackage[hyphens]{url}  % DO NOT CHANGE THIS
\usepackage{graphicx} % DO NOT CHANGE THIS
\usepackage{natbib}  % DO NOT CHANGE THIS AND DO NOT ADD ANY OPTIONS TO IT
\usepackage{caption} % DO NOT CHANGE THIS AND DO NOT ADD ANY OPTIONS TO IT
\usepackage{algorithm}
\usepackage{newfloat}
\usepackage{listings}
\DeclareCaptionStyle{ruled}{labelfont=normalfont,labelsep=colon,strut=off} % DO NOT CHANGE THIS
\floatstyle{ruled}
\newfloat{listing}{tb}{lst}{}
\floatname{listing}{Listing}
\usepackage{xcolor}
\usepackage{amsmath}
\usepackage{amssymb}
\usepackage{mathtools}
\usepackage{amsthm}
\definecolor{cvprblue}{rgb}{0.21,0.49,0.74}
\definecolor{darkgreen}{rgb}{0.18,0.60,0.24}
\definecolor{darkred}{rgb}  {0.86,0.13,0.02}
\definecolor{darkblue}{rgb} {0.33,0.31,0.77}
\usepackage{enumitem}
\usepackage[ruled,vlined,algo2e,noend]{algorithm2e}
\usepackage{graphicx}
\usepackage{multirow}
\usepackage{booktabs} % For formal tables
\usepackage{siunitx} % For table alignment
\usepackage{algpseudocode}

\makeatletter
\newcommand*\bigcdot{\mathpalette\bigcdot@{.5}}
\newcommand*\bigcdot@[2]{\mathbin{\vcenter{\hbox{\scalebox{#2}{$\m@th#1\bullet$}}}}}
\makeatother

\nocopyright
\title{Graph Structure Refinement with Energy-based Contrastive Learning}
\author{
    Xianlin Zeng\textsuperscript{\rm 1, 2},
    Yufeng Wang\textsuperscript{\rm 1},
    Yuqi Sun\textsuperscript{\rm 1},
    Guodong Guo\textsuperscript{\rm 3},
    Wenrui Ding\textsuperscript{\rm 1},
    Baochang Zhang\textsuperscript{\rm 1}
}
\affiliations{
    \textsuperscript{\rm 1}Beihang University, Beijing, P.R.China\\
    
     \textsuperscript{\rm 2}Postdoctoral Research Station at China RongTong Academy of Sciences Group Corporation Limited, Beijing, P.R.China\\
    
    \textsuperscript{\rm 3}Ningbo Institute of Digital Twin, Eastern Institute of Technology, Ningbo, P.R.China\\
    zengxianlin@buaa.edu.cn, wyfeng@buaa.edu.cn, sunyuqi@buaa.edu.cn, guodong.guo@mail.wvu.edu, \\ding@buaa.edu.cn, bczhang@buaa.edu.cn
    }

\usepackage{bibentry}
\begin{document}

\maketitle

\begin{abstract}
Graph Neural Networks (GNNs) have recently gained widespread attention as a successful tool for analyzing graph-structured data. However, imperfect graph structure with noisy links lacks enough robustness and may damage graph representations, therefore limiting the GNNs' performance in practical tasks. Moreover, existing generative architectures fail to fit discriminative graph-related tasks. To tackle these issues, we introduce an unsupervised method based on a joint of generative training and discriminative training to learn graph structure and representation, aiming to improve the discriminative performance of generative models. We propose an Energy-based Contrastive Learning (ECL) guided Graph Structure Refinement (GSR) framework, denoted as ECL-GSR. To our knowledge, this is the first work to combine energy-based models with contrastive learning for GSR. Specifically, we leverage ECL to approximate the joint distribution of sample pairs, which increases the similarity between representations of positive pairs while reducing the similarity between negative ones. Refined structure is produced by augmenting and removing edges according to the similarity metrics among node representations. Extensive experiments demonstrate that ECL-GSR outperforms \textit{the state-of-the-art on eight benchmark datasets} in node classification. ECL-GSR achieves \textit{faster training with fewer samples and memories} against the leading baseline, highlighting its simplicity and efficiency in downstream tasks.
\end{abstract}


\section{Introduction}
With the explosive growth of graph-structured data, Graph Neural Networks (GNNs) \cite{zhu2019robust, zhang2020gnnguard, zhang2019heterogeneous, xu2018powerful} have emerged as a potent deep learning tool, experiencing notable advancements across diverse applications, such as node classification \cite{velivckovic2017graph, kipf2016semi}, node clustering \cite{wang2019attributed, zhang2019attributed}, graph classification \cite{duvenaud2015convolutional, lee2019self}, link prediction \cite{peng2020graph, srinivasan2019equivalence}, recommendation systems \cite{wang2019sequential, yu2021self}, drug discovery \cite{wang2021multi}, and anomaly detection \cite{ding2019deep}. GNNs usually adopt a message-passing scheme \cite{gilmer2017neural}, aggregating information from neighboring nodes within the observed topology to compute graph representations.
%
%As commonly recognized, 
%
The strong representational capacity of most GNNs hinges on the assumption that graph structure is sufficiently reliable and perfectly noise-free \cite{szegedy2013intriguing}, considered as ground-truth information for model training. However, this assumption may hardly hold in real-world applications. 
This is due to the fact: i) Raw structure is typically derived from complex interactive systems, leading to inherent uncertainties and incomplete connections. Even worse, the GNN iterative mechanism with cascading effects repeatedly aggregates neighborhood features. Minor noises in the graph can propagate to adjacent nodes, influencing other node embeddings and potentially introducing further inaccuracies \cite{dai2018adversarial}. 
ii) Graph representation containing explicit structure is not informative enough to improve task performance. Raw topology only incorporates necessary physical connections, such as chemical bonds in molecules, and fails to capture abstract or implicit links among nodes. Furthermore, in various graph-related tasks, such as text graph in natural language processing \cite{yao2019graph} or scene graph for images in computer vision \cite{suhail2021energy}, the explicit structure may either be absent or unavailable.
%To fit the true connectivity distribution ptrue(v|vc) of vertices connected to target vertex vc, GraphGAN models the connectivity probability among vertices in a graph with a generator, G(v|vc;θG), to gener- ate vertices that are most likely connected to vc. 
%discriminative paradigms aim to learn aclassifier for predicting the existence of edges directly
%A discriminator D(v,vc;θD) outputs the edge probability between v and vc to differentiate the vertex pair generated by the generator from the ground truth.
%aims to discriminate the connectivity for the vertex pair (v, vc). D() outputs a single scalar representing the probability of an edge existing between v and vc

To tackle these challenges mentioned above, Graph Structure Refinement (GSR) involves learning invariant underlying relationships by extracting general knowledge from graph data, rather than relying solely on task-specific information. Therefore, the primary concern of GSR lies in graph representation learning, which can be broadly classified into two categories \cite{wang2018graphgan}. 
On one hand, graph generative models \cite{grover2016node2vec,dong2017metapath2vec,zheng2020distribution} assume that each node follows an inherent connectivity distribution. Edges are viewed as samples from these distributions, with the models enhancing node representations by optimizing the likelihood of these observed edges.
However, most downstream tasks are inherently discriminative, such as node classification and graph prediction. The state-of-the-art generative models have significantly deviated from discriminative architectures \cite{grathwohl2019your}. 
On the other hand, discriminative models \cite{velickovic2019deep,wang2018shine,hassani2020contrastive} focus on learning a classifier to predict the presence of edges directly. They output a single scalar to represent the probability between node pair, thereby differentiating the connectivity of edges.
Nonetheless, these models may suffer from overfitting to the training data, capturing noise instead of extracting latent useful features, as well as lack the ability to generalize across different datasets and diverse graph structures.
Recently, \cite{wang2022unified} and \cite{kim2022energy} establish a crucial connection between discriminative paradigms and Energy-based Models (EBMs) \cite{lecun2006tutorial}, creating a unified framework to generate higher-quality samples and better visual representations.
Motivated by these findings, we advocate for incorporating EBMs and Contrastive Learning (CL) to unlock the potential of generative models in addressing discriminative problems of graph-related tasks. 

%Contrastive Learning (CL) as a discriminative method trains neural networks by maximizing the agreement between different augmentations of the same data, such as SimCLR \cite{chen2020simple}, MoCo \cite{he2020momentum}, etc. %
%While Energy-based Models (EBMs) \cite{lecun2006tutorial} are generative methods aiming to learn an energy function $E_\theta(\chi)$ that assigns low energy values to inputs $\chi$ by directly maximizing the log-likelihood of the joint distribution \cite{grathwohl2019your}. 
%Both have effectively addressed numerous downstream issues, with EBMs excelling in image or text generation \cite{du2019implicit, qin2022cold}, and CL proving effective for large language model \cite{gunel2020supervised} and video question answering \cite{zhang2022erm}.
%As EBMs and CL are model-agnostic, they can flexibly fit the joint density of data points and labels. 

%\begin{figure}[t]
%\centering
%\includegraphics[width=.5\textwidth]{figure1.png} 
%\caption{Overview of the proposed ECL-GSR framework, comprising four essential components: data input, representation learning, structure refinement, and task evaluation.}
%\label{fig1}
%\end{figure}

In this paper, we explore a novel Energy-based Contrastive Learning (ECL) approach to guide the GSR framework, termed ECL-GSR, which integrates EBMs with CL for unsupervised graph representation learning. 
Specifically, ECL complements discriminative training loss with generative loss, supplying higher quality and more robust representations for downstream tasks.
Theoretically, we demonstrate that the existing discriminative loss is merely a specific instance of the ECL loss when the generative term is disabled.
Empirically, ECL can be interpreted as maximizing the joint log-likelihood of the similarity between positive sample pairs with EBMs and minimizing the similarity between negative ones with CL, indicating the augmentations of identical and different samples, respectively. 
In GSR, we perform edge prediction by adding or removing links based on the similarity probabilities among node representations, further refining the raw structure. 
Finally, we evaluate ECL-GSR on the node classification task using the refined graph. 
The major contributions are threefold as follows:
\begin{itemize}[leftmargin=*]
\item We present a novel ECL-GSR framework for joint graph structure and representation learning. It is the first work to combine EBMs with CL as generative and discriminative paradigms for GSR.
\item Contrary to most GSR methods, ECL-GSR is a straightforward implementation, demanding fewer training iterations, memory costs, and data samples to obtain the equivalent or better performance. %Its simple implementation is beneficial for the downstream task.
\item Extensive experiments on eight benchmark datasets demonstrate the superiority of ECL-GSR over current state-of-the-art methods. Ablation studies further confirm its effectiveness, efficiency, and robustness.
\end{itemize}

\section{Background}
\label{sec:background}
\subsection{Graph Structure Refinement}
Given a raw graph $G=(\mathcal{V},\mathcal{E},X)=(A,X)$ with noisy topology, where $\mathcal{V}$ is the set of $V=|\mathcal{V}|$ nodes, $\mathcal{E}$ is the set of $M=|\mathcal{E}|$ edges, $X\in\mathbb{R}^{V\times D}$ is the node feature matrix (the $i^{th}$ entry $x^i\in\mathbb{R}^D$ represents the attribute of node $v_i$), and $A\in\mathbb{R}^{V\times V}$ is the adjacency matrix ($A_{i,j}>0$ indicates ${e_{i,j}=(v}_i,v_j),{i,j}\in{M}$). The target of graph structure refinement \cite{zhu2021survey} is to acquire a refined graph $\widetilde{G}$ with a clean adjacency $\widetilde{A}$, along with corresponding representation $\widetilde{Z}\in\mathbb{R}^{V\times\widetilde{F}}$, $\widetilde{F}\ll D$, for downstream tasks. %low-dimensional with noisy topology

\subsection{Energy-based Models}
Given a point $\chi$ sampled from the data distribution $p_d(\chi)$, EBMs assign a scalar-valued energy function $E_\theta(\chi)\in\mathbb{R}$ by a DNN with parameters $\theta$. The energy function define a probability distribution using the Boltzmann distribution $p_\theta(\chi)=\frac{\exp({-E}_\theta(\chi))}{Z(\theta)}$, where $Z(\theta)$ is a normalizing constant or partition function ensuring $p_\theta$ integrates to 1. EBMs leverage the defined distribution $p_\theta$ to model the data distribution $p_d$ by minimizing the negative log-likelihood of $p_\theta$ under $p_d$, as indicated by:
\begin{equation}
\label{eq1}
\mathop{\min}\limits_\theta\mathbb{E}_{\chi\sim p_d}[-\log{p_\theta(\chi)}]. 
\end{equation}
The derivative of the negative log-likelihood $\mathcal{L}(\theta)$ is:
\begin{equation}
 \label{eq2}
 %\begin{aligned}
\nabla_\theta\mathcal{L}(\theta)\cong\mathbb{E}_{{\mathcal{\chi}^+
\sim p}_d}[\nabla_\theta E_\theta(\mathcal{\chi}^+)]\\
-\mathbb{E}_{\mathcal{\chi}^-\sim p_\theta}[\nabla_\theta E_\theta(\mathcal{\chi}^-)] \, .
 %\end{aligned}
 \end{equation}
Eq. \ref{eq2} decreases the energy values of positive samples $\chi^+$ while increasing those of negative ones $\chi^-$ \cite{hinton2002training}. However, computing $Z(\theta)$ for most parameterizations of $E_\theta(\mathcal{\chi})$ is intractable. We employ Stochastic Gradient Langevin Dynamics (SGLD) \cite{welling2011bayesian} derived from Markov Chain Monte Carlo (MCMC) methods to reduce the mixing time of sampling procedure. Specifically, it generates $p_\theta$ as an approximation of $p_d$ via iteratively updating $\chi$, denoted as:
\begin{equation}
 \label{eq3}
\chi_{k+1}=\chi_k-\frac{\lambda}{2}{\nabla_\chi}E_\theta(\chi_k)+\omega_k \, ,
 \end{equation}
\noindent where $\omega_k\sim\mathcal{N}(0,\lambda)$. As $k\rightarrow\infty$ and $\lambda\rightarrow0$, then $p_\theta$ converges to $p_d$. This process generates data samples through the energy function implicitly rather than explicitly.

\subsection{Contrastive Learning}
Given a set of random variables $\{\chi_n\}_{n=1}^N$, we define a data augmentation $\mathcal{T}$ to generate two distinct views $\nu_n=t(\chi_n)$,$\ \nu_n^\prime=t^\prime(\chi_n)$, i.e., $t,t^\prime\sim\mathcal{T}$. CL constitutes an unsupervised framework for representation learning, aiming to maximize the mutual information $I$ between the representations of two views $\nu_n$ and $\nu_m^\prime$ w.r.t the joint distribution $p(\nu_n,\ \nu_m^\prime)$. This is expressed as: 
\begin{equation}
\label{eq4}
\mathop{\max}\limits_{\mathcal{D}_\theta}{I(z_n,\ z_m^\prime)}\, ,
\end{equation}
\noindent where $z_n=\mathcal{D}_\theta(\nu_n)$ is the representation and $\mathcal{D}_{\theta}(\bigcdot)$ is a parametric DNN. When $n=m$, the views $(\nu_n,\nu_m^\prime)$ are referred to as a positive pair with the same marginal distribution. Conversely, they are called a negative pair. In practice, each pair provides supervisory information to the other, playing a role similar to that of labels in a supervised manner. CL trains $\mathcal{D}_\theta$ to encourage $z_n$ and its positive pair $z_n^\prime$ to be close in the projection space while pushing away representations of all negative pairs $z_m^\prime$. This principle has been proven to be key in boosting performance \cite{chen2020simple}.
%\subsection{Motivation}
%Currently, EBMs as generative models have effectively tackled numerous downstream issues, such as image generation and missing data imputation. However, as Grathwohl et al. \cite{grathwohl2019your} point out, most downstream tasks are inherently discriminative, and state-of-the-art generative models have significantly deviated from discriminative architectures.
%Encouragingly, Wang et al. \cite{wang2022unified} and Kim et al. \cite{kim2022energy} have established a crucial connection between discriminative models and EBMs, creating a unified framework to generate higher-quality samples and develop meaningful visual representations.

%In this paper, a substantial gap exists between the generative GSR and the discriminative node classification tasks. We advocate for the incorporation of EBMs and CL paradigms to unlock the potential of generative models in addressing downstream discriminative problems. As EBMs and CL are model-agnostic, they offer the flexibility to define and fit the joint density of data points and labels. We believe that they can serve as a catalyst for advancing research in graph representation learning. 

\begin{figure*}[ht]
\centering
\includegraphics[width=1.\textwidth]{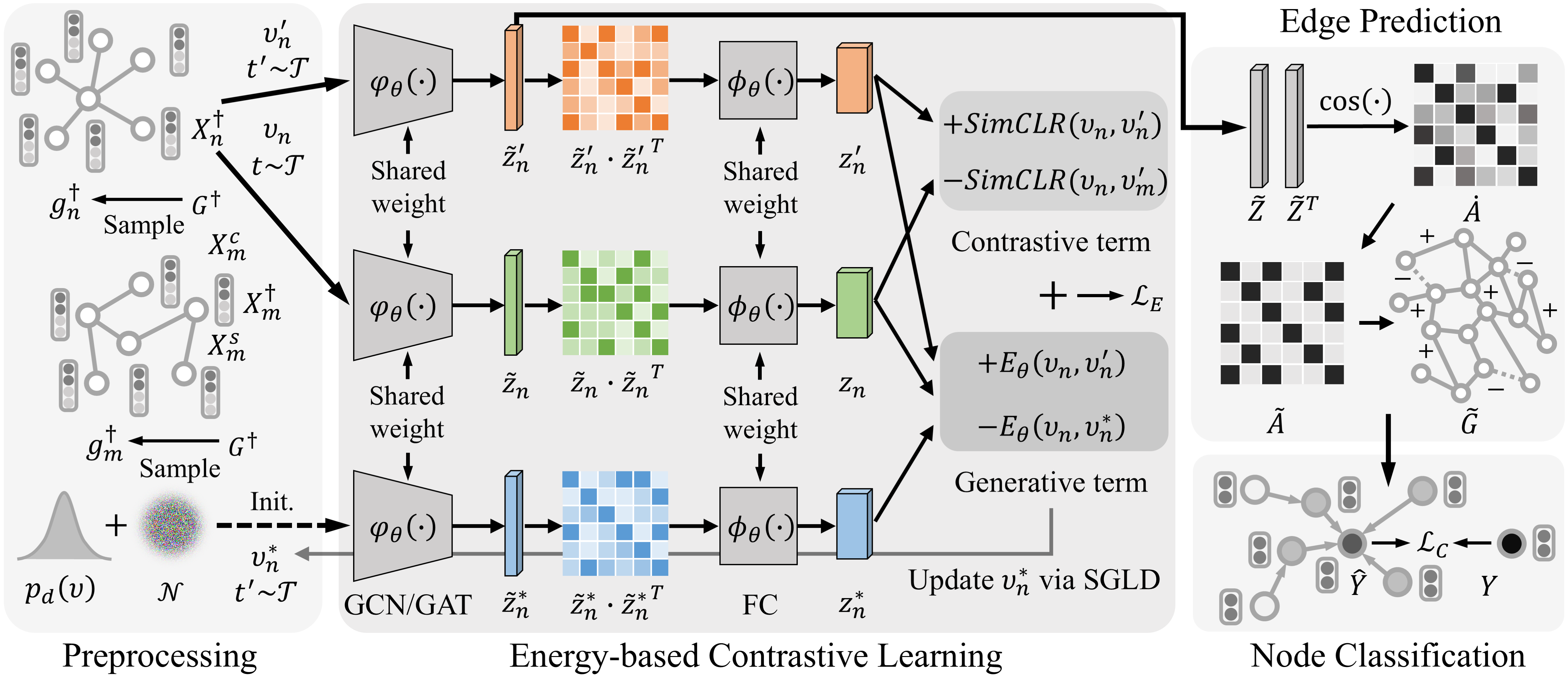} % Reduce the figure size so that it is slightly narrower than the column.
\caption{Illustration of the procedure of ECL-GSR. Preprocessed dual-attribute graph undergoes data augmentations, energy-based contrastive learning, and edge prediction, achieving structure refinement. Refined graph is applied in node classification.}
\label{fig2}
\end{figure*}

\section{Methodology}
In this section, we delineate the proposed ECL-GSR framework. As shown in Fig.~\ref{fig2}, our pipeline consists of four steps: preprocessing, energy-based contrastive learning, edge prediction, and node classification. Initially, we construct a dual-attribute graph by extracting contextual and structural information and acquiring subgraphs as input through edge sampling. Then, the ECL approach is introduced with the theorem and implementation. Next, we fine-tune the graph structure and evaluate the node classification task with raw features. Lastly, we present the final training objective. 
\subsection{Preprocessing}
\paragraph{\textbf{Dual-attribute graph}}
Our framework can make full use of all trustworthy observations to maximize informativeness by constructing a dual-attribute graph. We concatenate the contextual information $X^c$, and structural embedding $X^s$ as a new attribute, where $X^c$ is derived directly from the raw node features and $X^s$ is extracted using the DeepWalk \cite{perozzi2014deepwalk}. Finally, the dual-attribute graph is represented as $G^\dagger=(X^\dagger,A)$, where $X^\dagger=[X^c,X^s]$ is the new node features.
\paragraph{\textbf{Edge sampling}}
To address memory limitations, neighbor sampling techniques enable stochastic training on large graphs $G$ by decomposing them into smaller subgraphs $g$. Each subgraph sequentially contributes to GNNs’ optimization, executing multiple gradient descent steps within a training epoch. We independently select several edges (node pairs) from edge set $\left\{e_{i,j}\right\}_{i,j=1}^M$ to produce a subgraph. This process yields a mini-batch of subgraphs $\left\{g_n\right\}_{n=1}^N$, each with a fixed number of edges.

\subsection{Energy-based Contrastive Learning}
Initially, we define $p_d$ as a distribution of graph data and $\mathcal{T}$ as a set of predetermined data augmentation operators. Given a dual-attribute subgraph $g^\dagger$ and two augmentation views $t,t^\prime\sim\mathcal{T}$ selected uniformly at random, we propose an ECL approach to build a joint distribution $p_\theta(\nu,\nu^\prime)$ over two views $\nu,\nu^\prime=t(g^\dagger),t^\prime(g^\dagger)$, aiming to approximate the data distribution $p_d(\nu,\nu^\prime)$.%, $\nu^\prime=t^\prime(g^\dagger)$. 

\noindent\textbf{Definition 1.} \textit{The joint distribution $p_\theta(\nu,\nu^\prime)$ can be defined as:
\begin{equation}
\label{eq5}
p_\theta(\nu,\nu^\prime)=\frac{\exp(-f_\theta(\nu,\nu^\prime))}{Z(\theta)},\, 
\end{equation}
\noindent where $Z(\theta)=\int\int\exp(-f_\theta(\nu,\nu^\prime))d\nu d\nu^\prime$.}

\noindent Building upon the assumption that semantically similar pairs $(\nu,\nu^\prime)$ have nearby projections with high $p_d$, while dissimilar ones would correspond to distant projections with low $p_d$, we solve for the distance between $\nu$ and $\nu^\prime$. Let $f_\theta(\bigcdot)=\phi_\theta(\varphi_\theta(\bigcdot))$, $\varphi_\theta(\bigcdot)$ is a GNN encoder, and $\phi_\theta(\bigcdot)$ is a linear projection. $z=f_\theta(\nu)$ is the corresponding representation. The term $\Vert z-z^\prime\Vert$ indicates the inverse of semantic similarity of $\nu$ and $\nu^\prime$. To approximate $p_\theta(\nu,\nu^\prime)$ to $p_d(\nu,\nu^\prime)$, Eq. \ref{eq1} can be rephrased as: 
\begin{equation}
\label{eq6}
\mathop{\min}\limits_\theta{\mathbb{E}_{p_d}[-\log{p_\theta(\nu,\nu^\prime)}]}\,.
\end{equation}

\noindent\textbf{Proposition 1.} \textit{The joint distribution $p_\theta(\nu,\nu^\prime)$ can be formulated as an EBM:}
\begin{equation}
\label{eq7}
p_\theta(\nu,\nu^\prime)=\frac{\exp(-E_\theta(\nu,\nu^\prime))}{Z(\theta)}, \,
\end{equation}
\textit{\noindent where $E_\theta(\nu,\nu^\prime)={\Vert z-z^\prime\Vert^2}/\tau$, and $\tau$ is a temperature parameter.}

\noindent The gradient of the objective of Eq. \ref{eq6} is expressed as:
\begin{equation}
\label{eq8}
\begin{aligned}
\nabla_\theta\mathbb{E}_{p_d}[-\log&{p_\theta(\nu,\nu^\prime)}]=\\\mathbb{E}_{p_d}&[\nabla_\theta E_\theta(\nu,\nu^\prime)]-\mathbb{E}_{p_\theta}[\nabla_\theta E_\theta(\nu,\nu^\prime)]\, .
\end{aligned}
\end{equation}
To avoid directly calculating $Z(\theta)$, we employ Bayes' rule \cite{bayes1763lii} to reformulate ${\mathbb{E}_{p_d}[-\log{p_\theta(\nu,\nu^\prime)}]}$ as:
\begin{equation}
\label{eq9}
\begin{aligned}
\mathbb{E}_{p_d}[-\log&{p_\theta(\nu,\nu^\prime)}]=\\\mathbb{E}&_{p_d}[-\log{p_\theta(\nu^\prime|\nu)}]+\mathbb{E}_{p_d}[-\log{p_\theta(\nu)}]\, ,
\end{aligned}
\end{equation}
\noindent where $p_\theta(\nu)$ is the marginal distribution of $p_\theta(\nu,\nu^\prime)$ over $\nu^\prime$. 

\noindent\textbf{Theorem 1.} \textit{The marginal distribution $p_\theta(\nu)$ is an EBM:}
\begin{equation}
\label{eq10}
p_\theta(\nu)=\frac{\exp(-E_\theta(\nu))}{Z(\theta)}, \,
\end{equation}
\textit{where $E_\theta(\nu)=-\log{\int{\exp{(-\Vert z-z^\prime\Vert^2/\tau})d\nu^\prime}}$.}

\noindent Its proof is detailed in Appendix A.1. The gradient of the objective of Eq. \ref{eq10} is defined as:
\begin{equation}
\label{eq11}
\nabla_\theta\mathbb{E}_{p_d}[-\log{p_\theta(\nu)}]=\mathbb{E}_{p_d}[\nabla_\theta E_\theta(\nu)]-\mathbb{E}_{p_\theta}[\nabla_\theta E_\theta(\nu)]\,.
\end{equation}
According to Eq. \ref{eq9}, the objective of ECL is decomposed into the generative term and discriminative term, given by:
\begin{equation}
\label{eq12}
\mathcal{L}_b(\theta)=\mathbb{E}_{p_d}[-\log{p_\theta(\nu^\prime|\nu)}]+{\alpha\mathbb{E}}_{p_d}[-\log{p_\theta(\nu)}]\, ,
\end{equation}
\noindent where $\alpha$ is a hyperparameter to trade off the strength of two terms. According to Eq. \ref{eq11}, the gradient of Eq. \ref{eq12} is written as:
\begin{equation}
\label{eq13}
\begin{aligned}
\nabla_\theta\mathcal{L}_{b}(\theta)=\mathbb{E}_{p_d}&[-\nabla_\theta\log{p_\theta(\nu^\prime|\nu)}]+\\&{\alpha\mathbb{E}}_{p_d}[\nabla_\theta E_\theta(\nu)]-{\alpha\mathbb{E}}_{p_\theta}[\nabla_\theta E_\theta(\nu)]\, .
\end{aligned}
\end{equation}
By this way, $Z(\theta)$ ingeniously cancels itself out in the discriminative term without additional calculations. For the generative term, we merely need to sample $\nu^\ast$ from $p_d(\nu)$ with adding Gaussian noise $\mathcal{N}(0,\lambda)$ and iteratively optimize $\nu^\ast$ through SGLD, as indicated in Eq. \ref{eq3}.

\noindent\textbf{Implementation 1.} \textit{To implement the training of ECL, we approximate the generative term and discriminative term of Eq. \ref{eq12}, respectively, using the empirical mean of $p_\theta(\nu)$.} 

\noindent Given a mini-batch of samples $\{(\nu_n,\nu_n^\prime)\}_{n=1}^N$, along with its representations $\{(z_n,z_n^\prime)\}_{n=1}^N$, we have $N$ positive and $2(N-1)$ negative samples. Therefore, the empirical mean ${\hat{p}}_\theta(\nu_n)$ \cite{kim2022energy} is defined as:
\begin{equation}
\label{eq14}
{\hat{p}}_\theta(\nu_n)=\frac{1}{2N}\sum_{\nu_m^\prime:\nu_m^\prime\neq\nu_n}^{2(N-1)}{p_\theta(\nu_n,\nu_m^\prime)}\, .
\end{equation}
For the discriminative term, we utilize $\frac{p_\theta(\nu_n,\nu_n^\prime)}{{\hat{p}}_\theta(\nu_n)}$ to approximate the conditional probability density $p_\theta(\nu^\prime|\nu)$. According the SimCLR framework \cite{chen2020simple}, ${\mathop{min}\limits_\theta\mathbb{E}_{p_d}}[-\log{\hat{p}_\theta(\nu_n^\prime|\nu_n)}]$ can be represented as:
\begin{equation}
\label{eq15}
\mathop{\min}\limits_{z\in f_\theta(\nu)}{-\log\left(\frac{\exp{(-\Vert z_n-z_n^\prime\Vert^2/\tau)}}{\frac{1}{2N}\sum_{\nu_m^\prime:\nu_m^\prime\neq\nu_n}^{2(N-1)}\exp{(-\Vert z_n-z_m^\prime\Vert^2/\tau)}}\right)}\, .
\end{equation}
Considering only $N$ positive samples in the generative term, we simplify Eq. \ref{eq14} to $\hat{p}_\theta(\nu_n)=\frac{1}{N}\sum_{n=1}^{N}{p_\theta(\nu_n,\nu_n^\prime)}$. The approximation of $\mathop{\min}\limits_\theta\mathbb{E}_{p_d}[-\log{{\hat{p}}_\theta({\nu}_n)}]$ is denoted as:
\begin{equation}
\label{eq16}
\mathop{\min}\limits_{z\in f_\theta(\nu)}{-\log\left(\sum_{n=1}^N\exp{(-\Vert z_n-z_n^\prime\Vert^2/\tau)}\right)}\, .
\end{equation}
In summation, the final objective of ECL is:
\begin{equation}
\label{eq17}
\mathcal{L}_{E}(\theta)=\mathcal{L}_b(\theta)+\beta\mathcal{L}_r(\theta)\, ,
\end{equation}
\noindent where $\mathcal{L}_r(\theta)=\frac{1}{2N}\sum_{n\neq m}{E_\theta\left(\nu_n,\nu_m^\prime\right)}^2$ is the $L_2$ regularization loss to prevent gradient overflow due to the excessive energy values. $\beta$ is also a trade-off hyperparameter.

\subsection{Edge Prediction}
Upon completion of the ECL training, we are able to fine-tune the graph structure through edge prediction. The edge predictor receives the graph representation and subsequently outputs an edge probability matrix, denoted as $\dot{A}$. Each element $\dot{A}_{i,j}$ symbolizes the predicted probability of an edge existing between the pair of nodes $(v_i,v_j)$:
\begin{equation}
\label{eq18}
{\dot{A}}_{i,j}=Norm{(cos{({\widetilde{z}}_i,{\widetilde{z}}_j)})}\, ,
\end{equation}
\noindent where ${\widetilde{z}}_i,{\widetilde{z}}_j\in\widetilde{Z}$, $\cos{(\bigcdot)}$ is the cosine similarity function, $\widetilde{Z}$ denotes the representation output by the encoder $\varphi_\theta(\bigcdot)$, and $Norm{(\bigcdot)}$ is a normalization function. For the purpose of end-to-end training, we binarize $\dot{A}$ with the relaxed Bernoulli sampling \cite{zhao2021data} on each edge to produce the final matrix $\widetilde{A}$. %For the purpose of end-to-end training, we employ a relaxed Bernoulli sampling procedure as a differentiable approximation. To binarize the relaxed sample, we apply Gumbel-Softmax distribution to ensure global forward and backward passes of the framework, whose gradients are directly passed to the relaxed samples via the reparameterization trick \cite{jang2016categorical}. Each element $\widetilde{A}_{i,j}$ is denoted as:
%\begin{equation}
%\label{eq15}
%\widetilde{A}_{i,j}=\frac{1}{1+\exp(-{(\log\dot{A}_{i,j}+\mathcal{G})}/\tau)}+\frac{1}{2}\, ,
%\end{equation}
%\noindent where $\tau$ is also a temperature, and $\mathcal{G}\sim Gumbel(0,\ 1)$ is the Gumbel random variable. 
%Through this step of effective denoising, the refined structure is more informative and robust.

\begin{algorithm}[t]
	%\small
 	\SetAlgoLined
	\SetKwInOut{Majorization}{Majorization}\SetKwInOut{Minimization}{Minimization}
	\SetKwData{set}{set}
	\SetKwInOut{Initialization}{Initialization}
	\SetKwInOut{Input}{Input}\SetKwInOut{Output}{Output}
	\KwIn {Dual-attribute graph $G^\dagger$ with node classification label $Y$, EBM $E_\theta$, augmentation operators $t,t^\prime$, batch size $N$, number of batches $B$, SGLD iterations $K$, and training epochs $P$}
	\KwOut {The predicted label $\hat{Y}$}
	Construct $G^\dagger$ from $G$, randomly initialize $E_\theta(\bigcdot)$ (including $\varphi_\theta(\bigcdot)$, $\phi_\theta(\bigcdot)$) and $C_\theta(\bigcdot)$ with $\alpha, \beta, \tau, \mu$;
	
	\For{ $p=1,2,\ldots , P$}{
	\For{ $b=1,2,\ldots , B$}{
		Batch $\{g_n^\dagger\}_{n=1}^N$ from $G^\dagger$;
		
		Build $p_d(\nu,\nu^\prime)$ over $\nu,\nu^\prime=t(g^\dagger),t^\prime(g^\dagger)$;
		
		Sample $\left\{\nu_n,\nu_n^\prime\right\}_{n=1}^N$ from $p_d(\nu,\nu^\prime)$;
		
		Calculate the discriminative term of $\mathcal{L}_{b}$ with Eq. \ref{eq15};
		
		Sample $\left\{\nu_n^\ast\right\}_{n=1}^N$ from $p_d(\nu)$ with $\mathcal{N}(0,\lambda)$; 
		
		\For{ $k=1,2,\ldots , K$}{
		Sample $\omega_k\sim\mathcal{N}(0,\lambda)$; 
		
		Update $\{\nu^\ast_{\text{n},\text{k+1}}\}_{n=1}^N$ from $\{\nu_{\text{n},\text{k}}^\ast\}_{n=1}^N$ with Eq. \ref{eq3};
		}
		Calculate the generative term of $\mathcal{L}_{b}$ with Eq. \ref{eq16};
		
        Calculate $\nabla_\theta\mathcal{L}_{b}$ with Eq. \ref{eq13} and $\mathcal{L}_E$ with Eq. \ref{eq17};
        
        Calculate $\dot{A}$ with Eq. \ref{eq18} and binarize $\dot{A}$ to yield $\widetilde{A}$;
        
        Predict $\hat{Y}$ with $C_\theta$ and calculate $\mathcal{L}_C$ with $Y$;
        
        Update $\theta_E$ and $\theta_C$ to minimize $\mathcal{L}$ with Eq. \ref{eq19};
    
	}}
	\caption{The entire process of ECL-GSR}
	\label{alg1}
\end{algorithm}

\subsection{Node Classification}
Using $\widetilde{A}$ and $X$ as inputs, we utilize a simple three-layer GNN as a node classifier $C_\theta$, which can be instantiated with GCN or GAT architecture. Node representations are defined as $H=C_\theta(\widetilde{A},X)$, and the predicted label $\hat{Y}$ aligns with the ground truth $Y$. For each node representation $h_i\in H$, $\hat{y}_i\in\hat{Y}$ is denoted as $Softmax{(h_i)}$. The node classification loss $\mathcal{L}_{C}(\theta)$ is the cross-entropy between $\hat{Y}$ and $Y$. 

\begin{table*}[ht]
\begin{center} 

\resizebox{\linewidth}{!}{ 
\begin{tabular}{ccccccccc}
\toprule  %添加表格头部粗线 
\textbf{Method} & \textbf{Cora} & \textbf{Citeseer} & \textbf{Cornell} & \textbf{Texas} & \textbf{Wisconsin} & \textbf{Actor} & \textbf{Pubmed} & \textbf{OGB-Arxiv}\\
\midrule[0.8pt]
%Edge Hom. & 0.81 & 0.74 & 0.80 & 0.65 & 0.12 & 0.06 & 0.18 & 0.22 \\
GCN         & 81.46 \small{\small{$\pm$ 0.58}} & 71.36 \small{$\pm$ 0.31} & 47.84 \small{$\pm$ 5.55} & 57.83 \small{$\pm$ 2.76} & 57.45 \small{$\pm$ 4.30} & 30.01 \small{$\pm$ 0.77}  & \textbf{\textcolor{darkblue}{79.18 \small{$\pm$ 0.29}}} & \textbf{\textcolor{darkgreen}{70.77 \small{$\pm$ 0.19}}} \\
GAT         & 81.41 \small{$\pm$ 0.77} & 70.69 \small{$\pm$ 0.58} & 46.22 \small{$\pm$ 6.33} & 54.05 \small{$\pm$ 7.35} & 57.65 \small{$\pm$ 7.75} & 28.91 \small{$\pm$ 0.83}  & 77.85 \small{$\pm$ 0.42} & \textbf{\textcolor{darkblue}{69.90 \small{$\pm$ 0.25}}} \\
LDS         & 83.01 \small{$\pm$ 0.41} & \textbf{\textcolor{darkgreen}{73.55 \small{$\pm$ 0.54}}} & 47.87 \small{$\pm$ 7.14} & 58.92 \small{$\pm$ 4.32} & 61.70 \small{$\pm$ 3.58} & \textbf{\textcolor{darkblue}{31.05 \small{$\pm$ 1.31}}}  & {OOM}            & {OOM}            \\
GEN         & 80.21 \small{$\pm$ 1.72} & 71.15 \small{$\pm$ 1.81} & \textbf{\textcolor{darkblue}{57.02 \small{$\pm$ 7.19}}} & \textbf{\textcolor{darkblue}{65.94 \small{$\pm$ 4.13}}} & \textbf{\textcolor{darkblue}{66.07 \small{$\pm$ 3.72}}} & 27.21 \small{$\pm$ 2.05}  & 78.91 \small{$\pm$ 0.69} & {OOM}            \\
SGSR        & 83.48 \small{$\pm$ 0.43} & 72.96 \small{$\pm$ 0.25} & 44.32 \small{$\pm$ 2.16} & 60.81 \small{$\pm$ 4.87} & 56.86 \small{$\pm$ 1.24} & 30.23 \small{$\pm$ 0.38}  & 78.09 \small{$\pm$ 0.53} & {OOM}            \\
GRCN        & \textbf{\textcolor{darkblue}{83.87 \small{$\pm$ 0.49}}} & 72.43 \small{$\pm$ 0.61} & 54.32 \small{$\pm$ 8.24} & 62.16 \small{$\pm$ 7.05} & 56.08 \small{$\pm$ 7.19} & 29.97 \small{$\pm$ 0.71}  & 78.92 \small{$\pm$ 0.39} & {OOM}            \\
IDGL        & \textbf{\textcolor{darkgreen}{83.88 \small{$\pm$ 0.42}}} & 72.20 \small{$\pm$ 1.18} & 50.00 \small{$\pm$ 8.98} & 62.43 \small{$\pm$ 6.09} & 59.41 \small{$\pm$ 4.11} & 28.16 \small{$\pm$ 1.41}  & {OOM}            & {OOM}            \\
GAuG-O      & 82.20 \small{$\pm$ 0.80} & 71.60 \small{$\pm$ 1.10} & 57.60 \small{$\pm$ 3.80} & 56.90 \small{$\pm$ 3.60} & 54.80 \small{$\pm$ 5.70} & 25.80 \small{$\pm$ 1.00}  & \textbf{\textcolor{darkgreen}{79.30 \small{$\pm$ 0.40}}} & {OOM}  \\
SUBLIME     & 83.40 \small{$\pm$ 0.42} & 72.30 \small{$\pm$ 1.09} & \textbf{\textcolor{darkgreen}{70.29 \small{$\pm$ 3.51}}} & \textbf{\textcolor{darkgreen}{70.21 \small{$\pm$ 2.32}}} & \textbf{\textcolor{darkgreen}{66.73 \small{$\pm$ 2.44}}} & 30.79 \small{$\pm$ 0.68}  & 73.80 \small{$\pm$ 0.60}  & 55.50 \small{$\pm$ 0.10}\\
ProGNN      & 80.30 \small{$\pm$ 0.57} & 68.51 \small{$\pm$ 0.52} & 54.05 \small{$\pm$ 6.16} & 48.37 \small{$\pm$ 8.75}& 62.54 \small{$\pm$ 7.56} & 22.35 \small{$\pm$ 0.88}  & 71.60 \small{$\pm$ 0.46} & {OOM}            \\
CoGSL       & 81.76 \small{$\pm$ 0.24} & \textbf{\textcolor{darkblue}{73.09 \small{$\pm$ 0.42}}} & 52.16 \small{$\pm$ 3.21} & 59.46 \small{$\pm$ 4.36} & 58.82 \small{$\pm$ 1.52} & \textbf{\textcolor{darkgreen}{32.95 \small{$\pm$ 1.20}}}  & {OOM}            & {OOM}            \\
STABLE      & 80.20 \small{$\pm$ 0.68} & 68.91 \small{$\pm$ 1.01} & 44.03 \small{$\pm$ 4.05} & 55.24 \small{$\pm$ 6.04} & 53.00 \small{$\pm$ 5.27} & 30.18 \small{$\pm$ 1.00}  & {OOM}            & {OOM}            \\
NodeFormer  & 80.28 \small{$\pm$ 0.82} & 71.31 \small{$\pm$ 0.98} & 42.70 \small{$\pm$ 5.51} & 58.92 \small{$\pm$ 4.32} & 48.43 \small{$\pm$ 7.02} & 25.51 \small{$\pm$ 1.17}  & 78.21 \small{$\pm$ 1.43} & 55.40 \small{$\pm$ 0.23} \\
\midrule[0.5pt]
ECL-GSR    & \textbf{\textcolor{darkred}{84.06 \small{$\pm$ 0.84}}} & \textbf{\textcolor{darkred}{73.70 \small{$\pm$ 0.75}}} & \textbf{\textcolor{darkred}{71.27 \small{$\pm$ 2.06}}} & \textbf{\textcolor{darkred}{72.97 \small{$\pm$ 3.39}}} & \textbf{\textcolor{darkred}{67.79 \small{$\pm$ 1.03}}} & \textbf{\textcolor{darkred}{33.71 \small{$\pm$ 0.96}}}  & \textbf{\textcolor{darkred}{80.91 \small{$\pm$ 1.12}}} & \textbf{\textcolor{darkred}{71.09 \small{$\pm$ 0.31}}} \\ 
\bottomrule 		
\end{tabular}
}
\end{center}
\caption{\label{table1}Node classification accuracy (mean(\%)±std) with the standard splits on various benchmark datasets. The top three results are highlighted in \textbf{\textcolor{darkred}{first best}}, \textbf{\textcolor{darkgreen}{second best}}, and \textbf{\textcolor{darkblue}{third best}}, respectively. "OOM" indicates out of memory.}  
\end{table*}

\subsection{Training Objective}
During the training process, we can efficiently compute the joint classification loss $\mathcal{L}_{C}(\theta)$ and ECL loss $\mathcal{L}_{E}(\theta)$ using gradient descent-based backpropagation techniques. The overall loss is:
\begin{equation}
\label{eq19}
\mathop{\min}\limits_{\theta_E,\theta_C}\mathcal{L}(\theta)=\mathcal{L}_{E}(\theta)+\mu\mathcal{L}_{C}(\theta)\, ,
\end{equation}
\noindent where $\theta_E$ and $\theta_C$ are parameters of $E_\theta(\bigcdot)$ and $C_\theta(\bigcdot)$, respectively. 
The pseudocode of ECL-GSR is illustrated in Algorithm \ref{alg1}. The training stability  is presented in Appendix A.6.
%Thanks to stochastic training, our framework easily realizes an acceptable space complexity. In parallel, adopting ECL delivers comparable outcomes with fewer training epochs, data samples, and memory costs. %We conduct a detailed analysis of the ablation study.

\section{Experiments}
We conduct comprehensive experiments to sequentially evaluate the proposed framework's effectiveness, complexity, and robustness, addressing five research questions: RQ1: How effective is ECL-GSR on the node classification task? RQ2: How efficient is ECL-GSR in terms of training time and space? RQ3: How do ECL architecture and its hyperparameters impact the performance of node-level representation learning? RQ4: How robust is ECL-GSR in the face of structural attacks or noises? RQ5: What kind of refined structure does ECL-GSR learn?
\subsection{Experimental Setups}
\paragraph{\textbf{Datasets}} For extensive comparison, we execute experiments on eight benchmark datasets: four citation networks (Cora, Citeseer \cite{sen2008collective}, Pubmed \cite{namata2012query}, and OGB-Arxiv \cite{hu2020open}), three webpage graphs (Cornell, Texas, and Wisconsin \cite{pei2020geom}), and one actor co-occurrence network (Actor \cite{tang2009social}). %Dataset statistics are detailed in the Appendix A.7.
\paragraph{\textbf{Baselines}} To corroborate the promising performance of ECL-GSR, we compare it against 13 GSR baseline methods. %for node classification accuracy on homogeneous graph datasets. 
There are two GNN baselines (GCN \cite{kipf2016semi} and GAT \cite{velivckovic2017graph}), three adjacency matrix direct-optimization methods (NodeFormer \cite{wu2022nodeformer}, STABLE \cite{li2022reliable}, and ProGNN \cite{jin2020graph}), four probability estimation techniques (GEN \cite{wang2021graph}, GAuG-O \cite{zhao2021data}, SGSR \cite{zhao2023self}, and LDS \cite{franceschi2019learning}), and four metric learning approaches (SUBLIME \cite{liu2022towards}, GRCN \cite{yu2021graph}, CoGSL \cite{liu2022compact}, and IDGL \cite{chen2020iterative}).
\paragraph{\textbf{Implementation details}} Our framework operates on an Ubuntu system with an NVIDIA GeForce 3090 GPU, employing PyTorch 1.12.1, DGL 1.1.0, and Python 3.9.16. All experiments are conducted using the reimplementation of GSLB \cite{li2023gslb}.
We maintain the dimensions of contextual $X^c$ and structural $X^s$ features equal to that of raw attribute. Subgraph sampling batch size $N$ is fixed at 64 for efficiency consideration.
In ECL, the backbone $f_\theta(\cdot)$ is divided into $\varphi_\theta(\cdot)$ for encoding, utilizing three GCN layers with the hidden and output dimension $\widetilde{F}$ of 128, and $\phi_\theta(\cdot)$ for projection, comprising two fully-connected layers with an output dimension $F$ of 128. The learned representation $\widetilde{Z}$ is produced by $\varphi_\theta(\bigcdot)$. Batch normalization is discarded when utilizing SGLD. The data augmentation operator $\mathcal{T}$ is a random Gaussian blur.
For node classification, classifier $C_\theta(\bigcdot)$ mirrors the architecture of $\varphi_\theta(\cdot)$. Our model's final hyperparameters are set as: $\alpha$=0.1, $\beta$=0.01, $\mu$=0.01, and $\tau$=0.1.
We adopt the Adam optimizer with an initial learning rate of 0.001, halving every 20 epochs. The epochs $P$ for Cora, Citeseer, Cornell, Texas, and Wisconsin are 40, and those for Actor, Pubmed, and OGB-Arxiv are 80. The number of SGLD's iterations $K$ only takes 3 steps.

\begin{table*}[ht]
\begin{center} 
 
\resizebox{\linewidth}{!}{ 
\begin{tabular}{ccccccccc}
\toprule
\multirow{2}{*}{\textbf{Method}} & \multicolumn{4}{c|}{\textbf{Cora}} & \multicolumn{4}{c}{\textbf{Citeseer}} \\
%\cmidrule{2-9}
%\cline{2-9}
    & {1\%} & {3\%} & {5\%} & \multicolumn{1}{c|}{10\%} & {1\%} & {3\%} & {5\%} & {10\%} \\
\midrule
GCN         & 59.31 \small{$\pm$ 0.29} & 77.14 \small{$\pm$ 0.21} & 80.73 \small{$\pm$ 0.63} & \multicolumn{1}{c|}{83.53 \small{$\pm$ 0.42}} & 60.64 \small{$\pm$ 1.07} & 67.60 \small{$\pm$ 0.47} & 70.05 \small{$\pm$ 0.54} & 74.38 \small{$\pm$ 0.27} \\
GAT         & 65.36 \small{$\pm$ 0.99} & 76.36 \small{$\pm$ 0.61} & 81.73 \small{$\pm$ 0.21} & \multicolumn{1}{c|}{83.92 \small{$\pm$ 0.42}} & 58.48 \small{$\pm$ 2.35} & 68.41 \small{$\pm$ 0.76} & 70.73 \small{$\pm$ 0.22} & \textbf{\textcolor{darkgreen}{74.54 \small{$\pm$ 0.14}}} \\
LDS         & 68.47 \small{$\pm$ 1.11} & 78.06 \small{$\pm$ 0.98} & 81.42 \small{$\pm$ 0.66} & \multicolumn{1}{c|}{83.87 \small{$\pm$ 0.48}} & 61.35 \small{$\pm$ 1.57} & 67.29 \small{$\pm$ 1.34} & 70.82 \small{$\pm$ 0.79} & 74.54 \small{$\pm$ 0.49} \\
IDGL        & \textbf{\textcolor{darkgreen}{70.83 \small{$\pm$ 1.21}}} & \textbf{\textcolor{darkgreen}{78.60 \small{$\pm$ 0.28}}} & \textbf{\textcolor{darkgreen}{83.82 \small{$\pm$ 0.28}}} & \multicolumn{1}{c|}{85.51 \small{$\pm$ 0.08}} & 60.61 \small{$\pm$ 1.32} & 64.34 \small{$\pm$ 1.61} & 69.39 \small{$\pm$ 1.24} & 74.19 \small{$\pm$ 0.58} \\
SGSR         & 55.11 \small{$\pm$ 0.43} & 77.32 \small{$\pm$ 0.17} & 83.51 \small{$\pm$ 0.22} & \multicolumn{1}{c|}{\textbf{\textcolor{darkgreen}{85.56 \small{$\pm$ 0.25}}}} & 54.28 \small{$\pm$ 0.47} & \textbf{\textcolor{darkgreen}{71.61 \small{$\pm$ 0.17}}} & \textbf{\textcolor{darkgreen}{{72.88 \small{$\pm$ 0.20}}}} & 74.31 \small{$\pm$ 0.24} \\
GRCN        & 68.38 \small{$\pm$ 2.10} & 75.24 \small{$\pm$ 1.06} & 79.16 \small{$\pm$ 0.82} & \multicolumn{1}{c|}{84.82 \small{$\pm$ 0.41}} & 59.06 \small{$\pm$ 1.80} & 66.17 \small{$\pm$ 0.75} & 72.11 \small{$\pm$ 0.56} & 74.49 \small{$\pm$ 0.73} \\
CoGSL       & 64.43 \small{$\pm$ 3.35} & 73.21 \small{$\pm$ 1.10} & 79.02 \small{$\pm$ 3.22} & \multicolumn{1}{c|}{81.05 \small{$\pm$ 0.53}} & 56.41 \small{$\pm$ 0.91} & 66.60 \small{$\pm$ 0.79} & 69.96 \small{$\pm$ 0.56} & 74.17 \small{$\pm$ 0.53} \\
ProGNN      & 70.32 \small{$\pm$ 1.16} & 75.93 \small{$\pm$ 0.78} & 81.35 \small{$\pm$ 0.68} & \multicolumn{1}{c|}{82.01 \small{$\pm$ 0.67}} & 56.77 \small{$\pm$ 0.88} & 70.34 \small{$\pm$ 0.66} & 70.67 \small{$\pm$ 0.79} & 74.23 \small{$\pm$ 0.36} \\
SUBLIME     & 65.94 \small{$\pm$ 4.90} & 73.37 \small{$\pm$ 0.78} & 79.14 \small{$\pm$ 0.26} & \multicolumn{1}{c|}{82.37 \small{$\pm$ 0.20}} & 57.85 \small{$\pm$ 1.64} & 67.67 \small{$\pm$ 0.84} & 70.53 \small{$\pm$ 0.16} & 71.47 \small{$\pm$ 0.08} \\
NodeFormer  & 67.11 \small{$\pm$ 1.07} & 75.87 \small{$\pm$ 0.79} & 82.05 \small{$\pm$ 0.67} & \multicolumn{1}{c|}{83.92 \small{$\pm$ 0.45}} & \textbf{\textcolor{darkgreen}{67.03 \small{$\pm$ 0.89}}} & 67.84 \small{$\pm$ 0.60} & 70.65 \small{$\pm$ 1.05} & 73.03 \small{$\pm$ 0.37} \\
\midrule[0.5pt] %NODE, CoGSL GRCN SUBLIME
ECL-GSR & \textbf{\textcolor{darkred}{72.33 \small{$\pm$ 0.39}}} & \textbf{\textcolor{darkred}{79.99 \small{$\pm$ 0.21}}} & \textbf{\textcolor{darkred}{84.30 \small{$\pm$ 0.18}}} & \multicolumn{1}{c|}{\textbf{\textcolor{darkred}{85.71 \small{$\pm$ 0.20}}}} & \textbf{\textcolor{darkred}{68.06 \small{$\pm$ 0.54}}} & \textbf{\textcolor{darkred}{72.18 \small{$\pm$ 0.15}}} & \textbf{\textcolor{darkred}{73.38 \small{$\pm$ 0.22}}} & \textbf{\textcolor{darkred}{74.90 \small{$\pm$ 0.23}}} \\
\bottomrule 		
\end{tabular}
}
\end{center}
\caption{\label{table2}Node classification accuracy (mean(\%)±std) with the different train ratios on Cora and Citeseer datasets. The top two results are highlighted in \textbf{\textcolor{darkred}{first best}} and \textbf{\textcolor{darkgreen}{second best}}, respectively.} 
\end{table*}
\begin{figure*}[t]
\centering
\includegraphics[width=1.\textwidth]{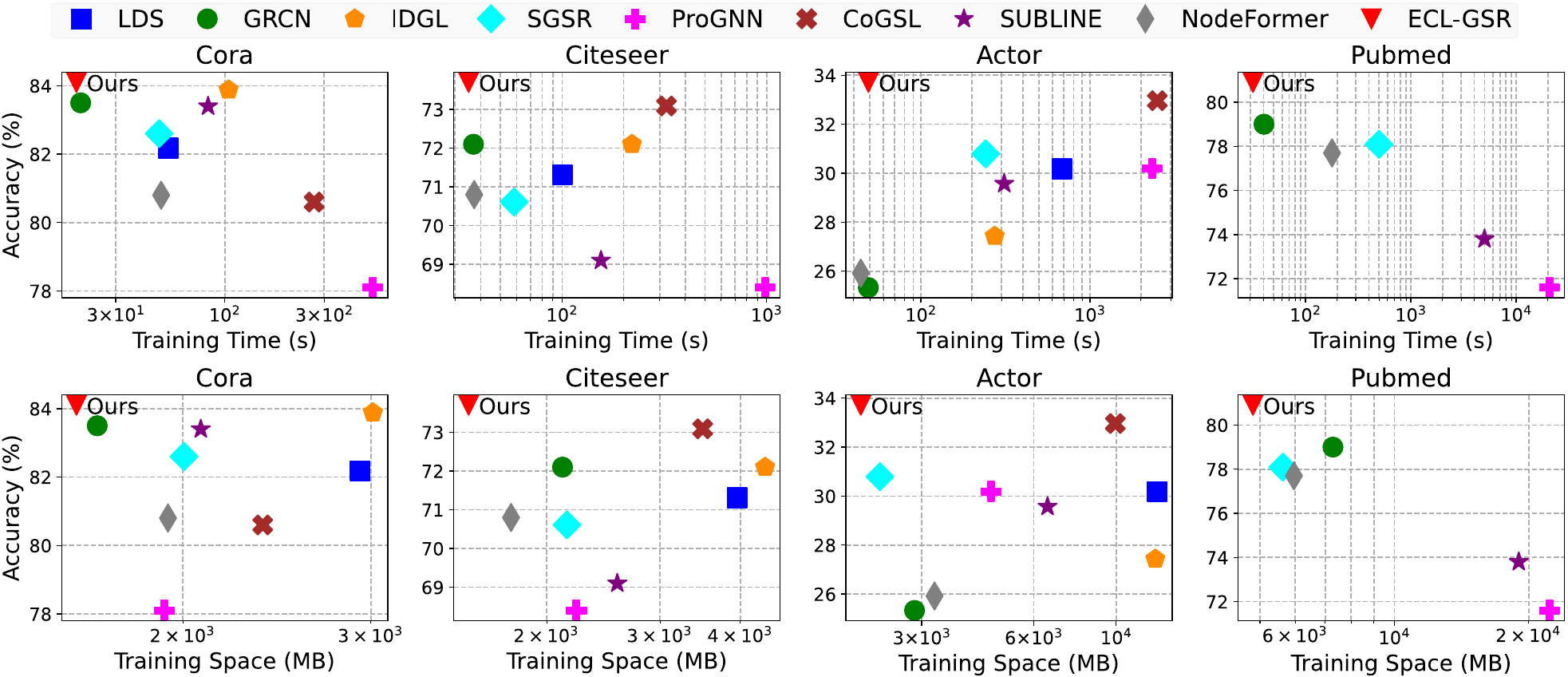} % Reduce the figure size so it is slightly narrower than the column.
\caption{Training time and space analysis on Cora, Citeseer, Actor, and Pubmed datasets.}
\label{fig3}
\end{figure*}
\subsection{Node Classification Performance (RQ1)}
\paragraph{\textbf{Evaluation on standard splits}}
As stated in Table \ref{table1}, three key observations can be made: 
i) ECL-GSR shows robust performance across all benchmark datasets, demonstrating its superior generalizability to diverse data. Notably, within the ambit of eight datasets, ECL-GSR achieves the state-of-the-art with margins ranging from 0.15\% to 1.61\% over the second-highest approach.
ii) Compared to other baselines, ECL-GSR exhibits enhanced performance stability and reduced standard deviation, particularly evident on the Cornell, Texas, Wisconsin, and Actor datasets.
iii) Whereas certain competing algorithms such as CoGSL, GEN, and IDGL encounter OOM errors with Pubmed and OGB-Arxiv, our approach achieves the state-of-the-art on large benchmarks.
\paragraph{\textbf{Evaluation on different train ratios}} 
%In Table \ref{table2}, the efficacy of our proposed ECL-GSR framework is markedly pronounced, mainly when operating under a paucity of supervised information. We meticulously orchestrated experiments at diminutive training ratios of 1\%, 3\%, 5\% to simulate scenarios with limited labeled data. The empirical outcomes unambiguously demonstrate that our framework excels, eclipsing the existing baselines in accuracy, particularly at lower training ratios.
%Our analysis corroborates the hypothesis that ECL-GSR not only achieves equivalent or superior results with a more constrained set of training samples but also retains a competitive edge when the availability of training data is significantly expanded. This dual advantage speaks volumes about the robustness and adaptability of ECL-GSR, making it a versatile tool in semi-supervised learning scenarios.
In Table \ref{table2}, we conduct experiments on Cora and Citeseer datasets with varying amounts of supervised information, specifically at training ratios of 1\%, 3\%, 5\%, and 10\%. The results indicate that our framework substantially outperforms existing baselines in terms of accuracy at a low training ratio. Among GSR approaches, we validate that ECL-GSR achieves equivalent or better performance with fewer training samples as well as maintains competitive performance at a high training ratio.
\begin{figure}[t]
\centering
\includegraphics[width=\linewidth]{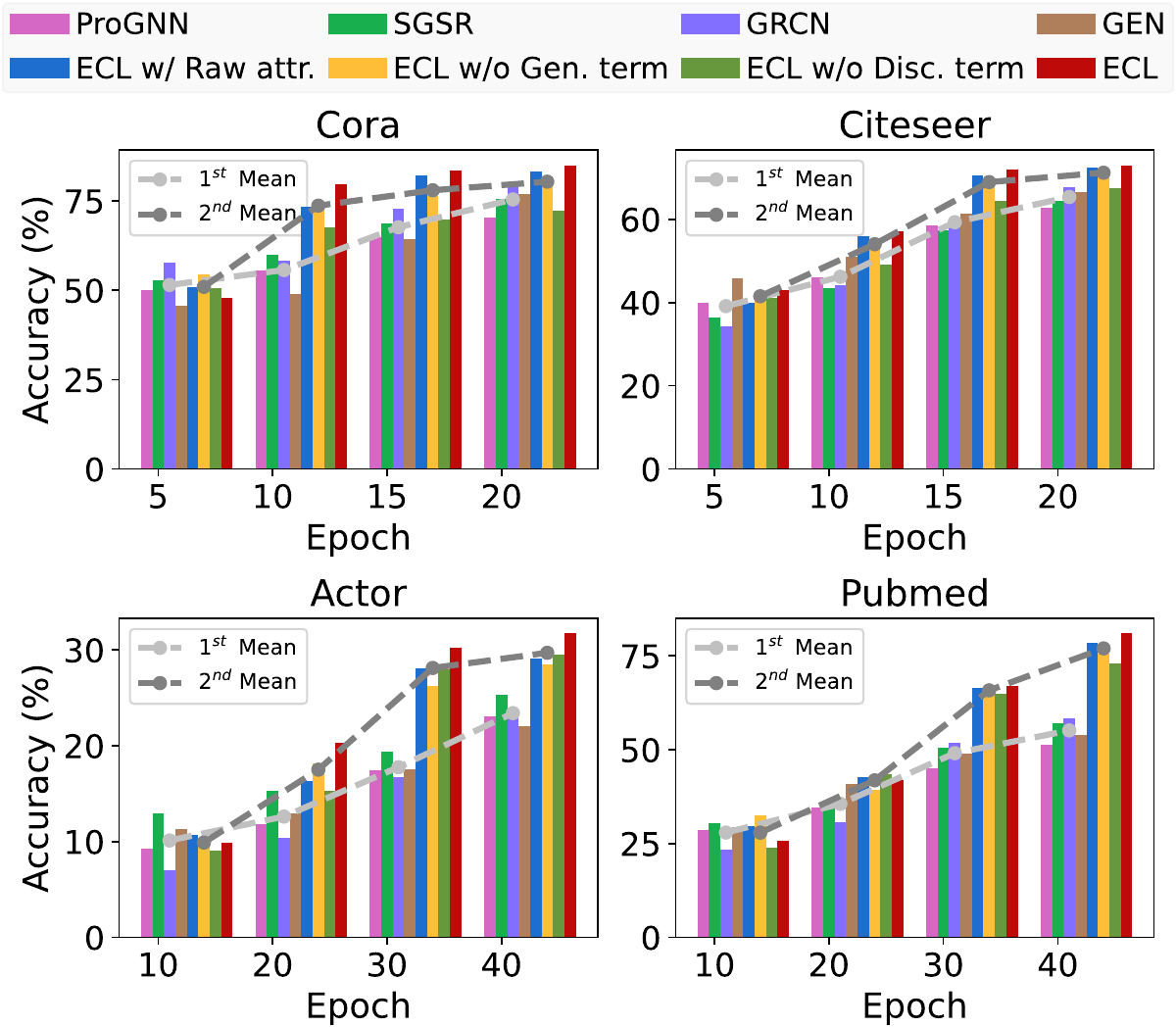}
\caption{Performance study of ECL-GSR variants and other baselines over multiple training epochs on four datasets.}
\label{fig4}
\end{figure}

\begin{figure}[t]
	\centering
	\includegraphics[width=\linewidth]{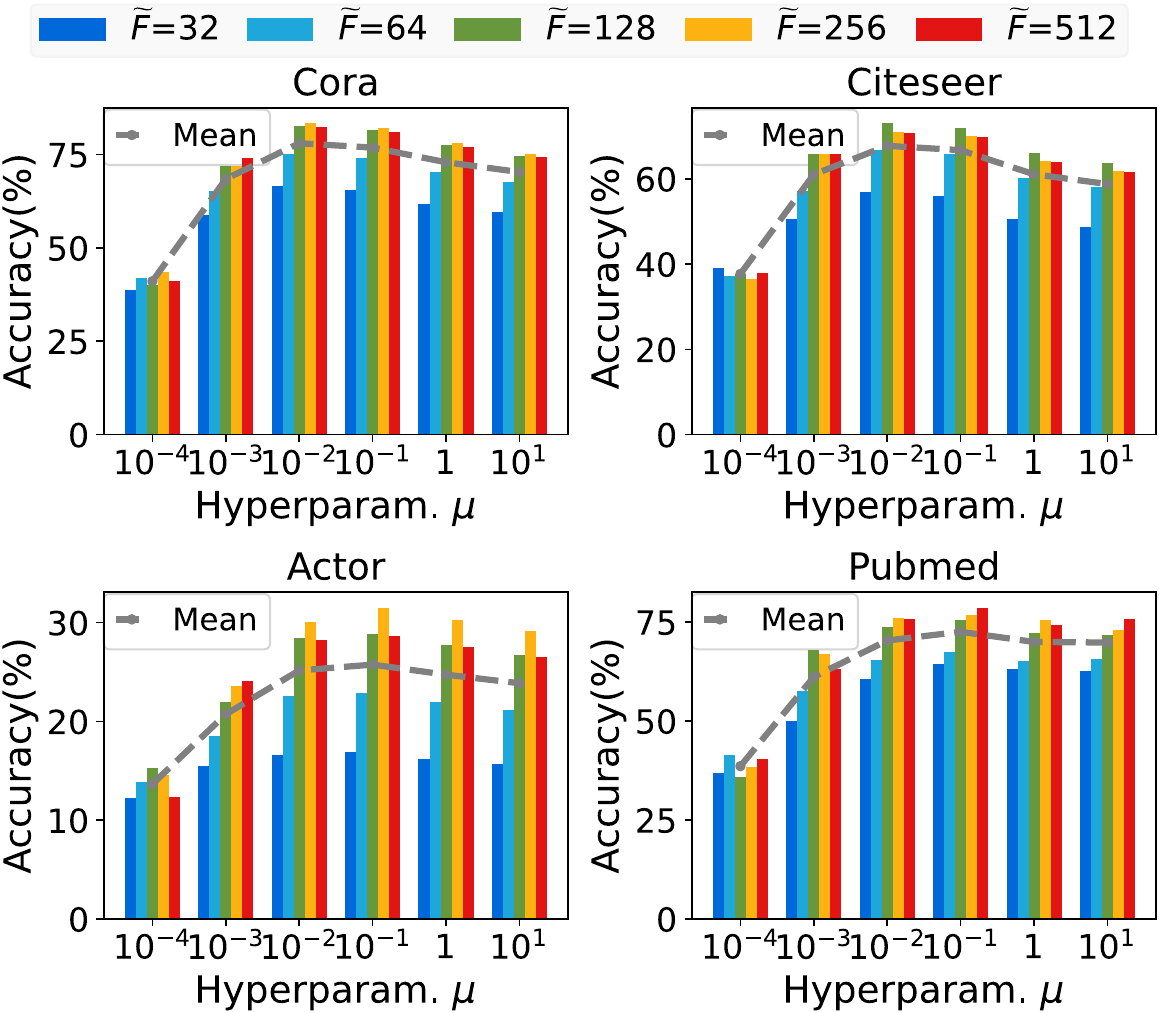}
    \caption{Hyperparameter $\mu$ and dimensionality $\widetilde{F}$ analysis of ECL-GSR on four datasets. “Mean” denotes the averages.}
      \label{fig5}
\end{figure}

\subsection{Efficiency and Scalability Analysis (RQ2)}
In this section, we analyze the efficiency and scalability of ECL-GSR on Cora, Citeseer, Actor, and Pubmed datasets. As illustrated in Fig.~\ref{fig3}, the position nearer to the figure's upper left corner signifies superior overall performance. For efficiency, the time complexity of performing an ECL-GSR is delineated by $\mathcal{O}(P\cdot K\cdot B)$, where $B$ represents the number of batches. The higher time efficiency of our approach stems from its expedited convergence, necessitating only a limited quantity of training epochs $P$ and iterations $K$. Regarding scalability, we can flexibly adapt the stochastic training by adjusting the mini-batch $N$, enabling to achieve an acceptable space complexity of $\mathcal{O}(N^2)$. 

Conventional GSR algorithms are typically hindered by their considerable time and space demands, constraining their applicability in large-scale graphs. Some alternative solutions, such as NodeFormer and SGSR, have been recognized for their speed, albeit due to diminished classification accuracy. Methods like CoGSL and LDS are notable for their effectiveness, yet they demand considerable computational and storage requirements. Conversely, ECL-GSR achieves advantages in terms of accuracy, speed, and memory usage, especially on Citeseer and Pubmed datasets.

\begin{figure}[H]
	\centering
	\includegraphics[width=\linewidth]{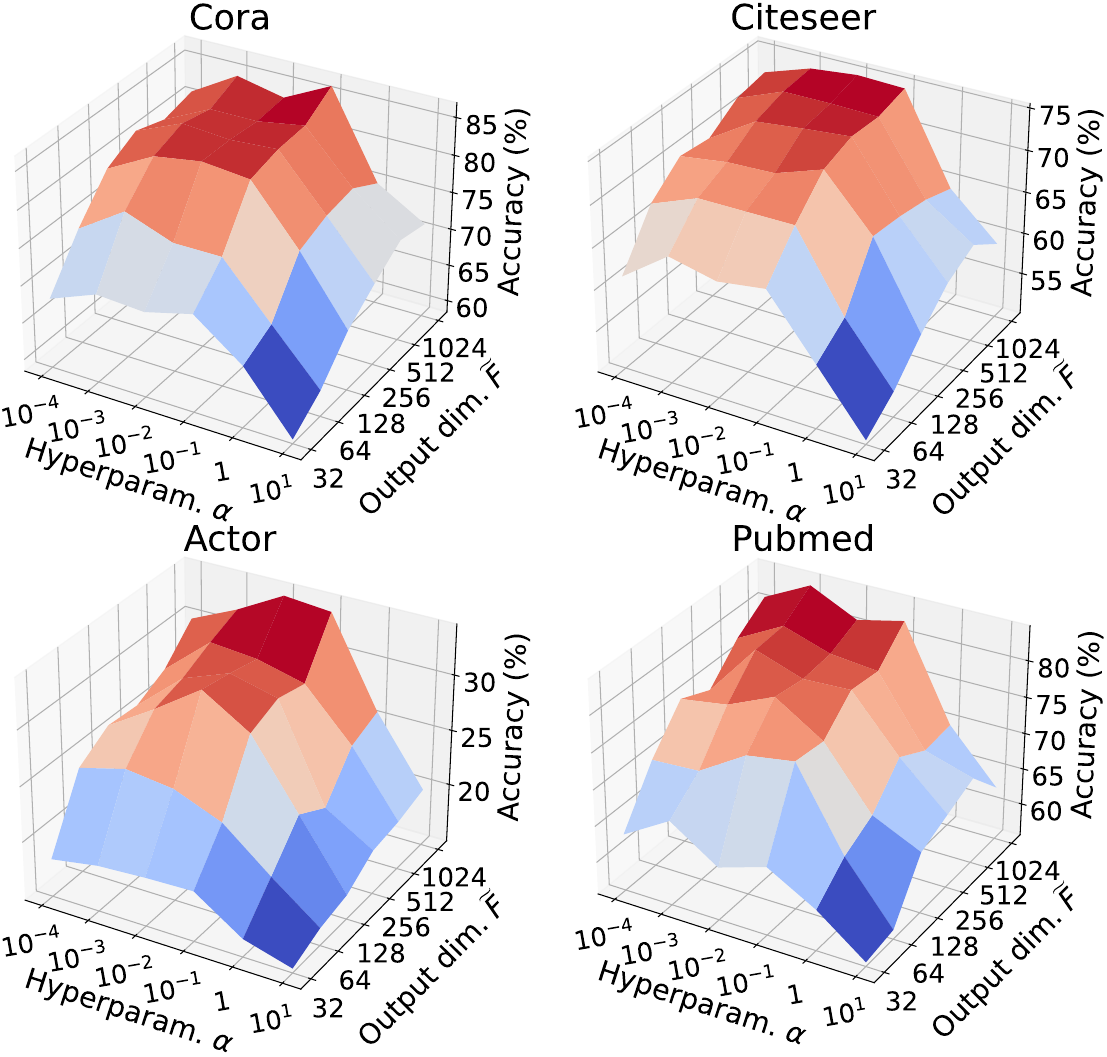}
    \caption{Hyperparameter $\alpha$ and dimensionality $\widetilde{F}$ analysis of ECL-GSR on four datasets.}
      \label{fig6}
\end{figure}  
\subsection{Ablation Study (RQ3)}
\paragraph{\textbf{Component analysis}} As illustrated in Fig.~\ref{fig4}, we investigate the impact of various configurations on Cora, Citeseer, Actor, and Pubmed datasets, evaluating the performance of GEN, GRCN, SGSR, ProGNN, ECL with raw attr., ECL without gen. term, ECL without disc. term, and full ECL over a range of training epochs.
“$1^{st}$ and $2^{nd}$ Mean” are averages of baselines and variations, respectively.
%Our findings indicate that: i) utilizing raw graph attributes without structural embeddings marginally reduces the accuracy of ECL-GSR, (ii) the notable decrease in performance when either generative and discriminative terms are omitted significantly underscores their vital roles in facilitating both intra- and inter-contrastive learning paradigms, and (iii) ECL-GSR and its variants surpass other baselines in achieving peak performance in fewer training epochs, highlighting our framework's swift convergence.
Our findings indicate that: i) Utilizing raw graph attributes without structural embeddings marginally reduces the accuracy of ECL-GSR. (ii) When either generative or discriminative terms are absent, a notable decrease in performance suggests a vital role for the combination of them. (iii) All variants reach their peak performance within fewer training epochs, highlighting our framework's swift convergence compared to other GSR.

\paragraph{\textbf{Parameter analysis}} The impacts of varying hyperparameter $\alpha$, $\mu$ and output dimension $\widetilde{F}$ on Cora, Citeseer, Actor, and Pubmed datasets are indicated in Fig.~\ref{fig5} and Fig.~\ref{fig6}, respectively.
%With regard to $\alpha$ on the x-axis, our analysis delves into the importance of the generative term relative to the discriminative term within the ECL framework. Empirical results advocate that an $\alpha$ setting exceeding 1.0 or higher yields suboptimal outcomes. The performance peaks at 0.1 and then experiences a marginal decline as $\alpha$ is decreased further.
%For the y-axis concerning $\widetilde{F}$, the output dimensionality is pivotal in striking a balance between sufficient representation capacity and the circumvention of overfitting. Diminished dimensions result in underperformance, attributable to inadequate representational space, whereas augmented dimensions uphold performance levels yet introduce additional computational complexity. We make an optimal trade-off between accuracy, hyperparameter, and dimensionality.
With respect to $\mu$, a low value weakens the constraint of classification loss, whereas a high value leads our framework to degrade to baseline, thereby diminishing the role of ECL.
Regarding $\alpha$, we explore the importance of the generative term relative to the discriminative term in ECL. It suggests that setting $\alpha$ to 1.0 or higher yields suboptimal results. The performance peaks at 0.1 and then experiences a marginal decline as $\alpha$ is decreased further. 
For selecting $\widetilde{F}$, it is crucial for balancing representational adequacy and preventing overfitting. Lower dimensions compromise performance due to insufficient representation, while higher dimensions maintain performance but add model complexity. 
%We make an optimal trade-off between accuracy, hyperparameter, and dimensionality.

\begin{figure}[t]
	\centering
	\includegraphics[width=\linewidth]{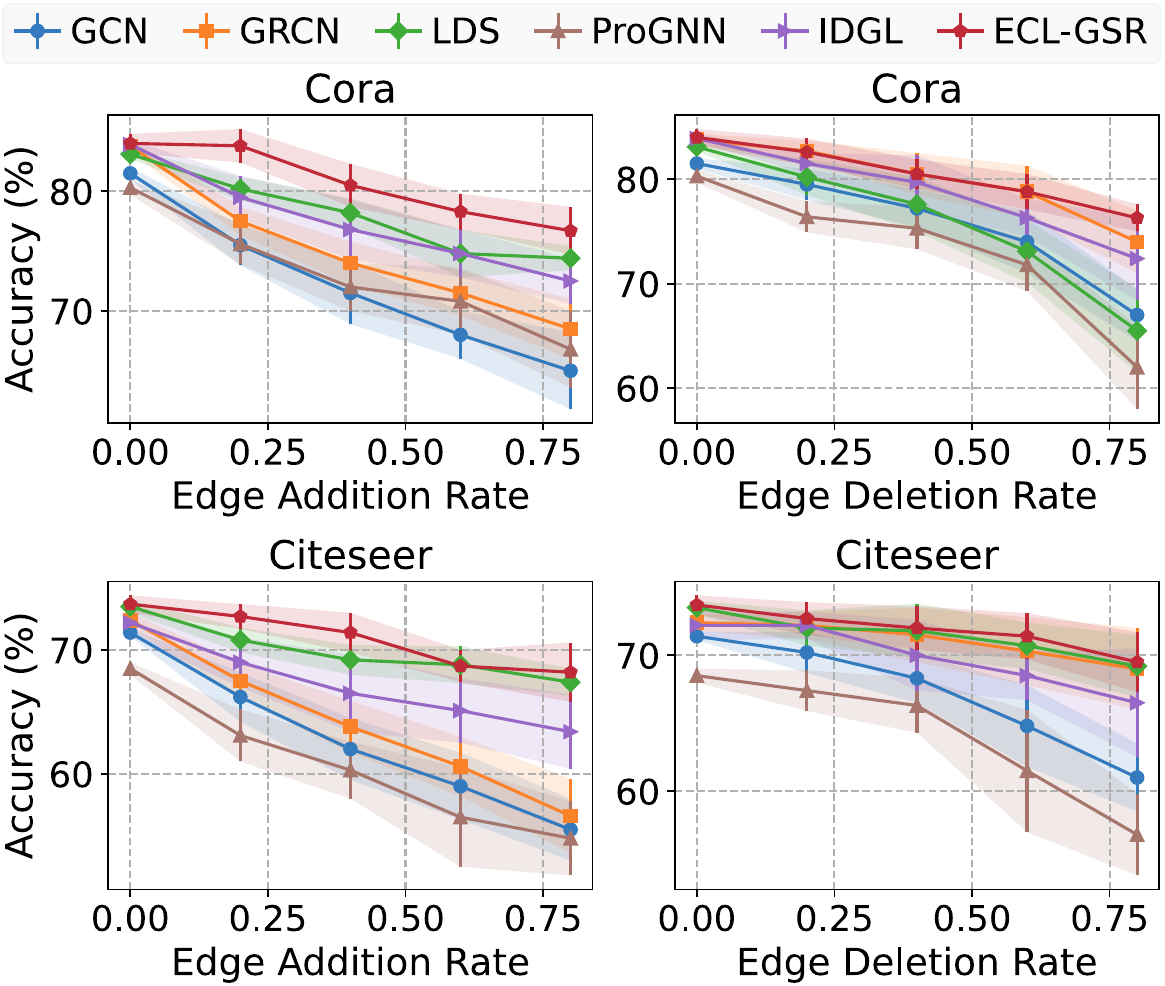} 
	\caption{Robustness analysis by randomly adding and removing edges on Cora and Citeseer datasets.}
	\label{fig7}
\end{figure}

\subsection{Robustness Analysis (RQ4)}
To evaluate the robustness of ECL-GSR, we randomly add or remove edges from the raw graph on Cora and Citeseer datasets and then evaluate the performance of various algorithms on the corrupted graph. We change the ratios of modified edges from 0 to 0.8 to simulate different noise intensities and compare our framework with GCN, GRCN, LDS, ProGNN, and IDGL. 
As revealed in Fig.~\ref{fig7}, the performance of models generally shows a downward trend with increased attack intensity. Nonetheless, GSR approaches commonly exhibit better stability than GCN baseline. When edge addition and deletion rates increase, ECL-GSR consistently achieves better or comparable results in both scenarios, indicating its robustness in severe structural attacks.

\subsection{Structure Visualization (RQ5)}
\begin{figure}[t]
\centering
\includegraphics[width=\linewidth]{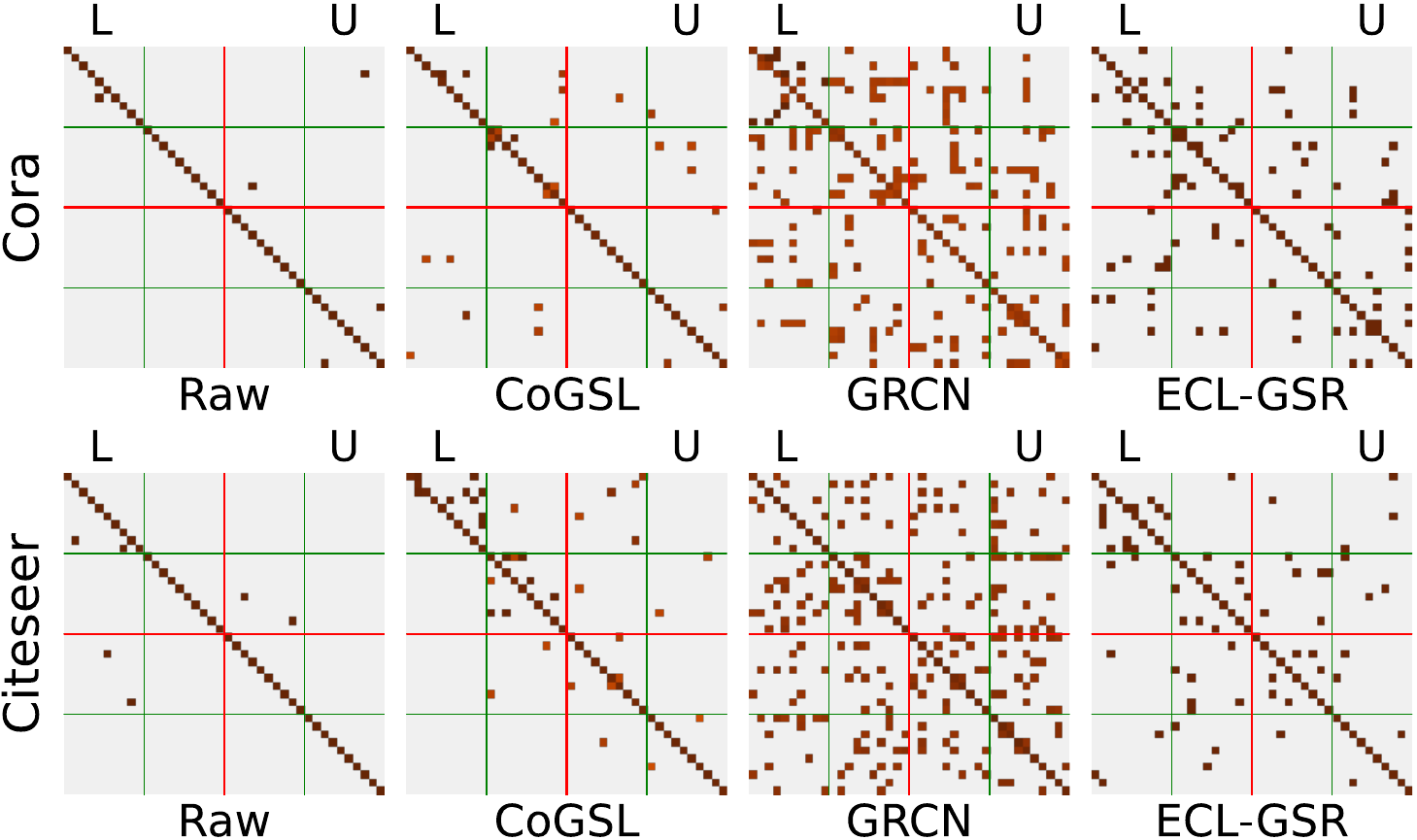} 
\caption{Visualization of the refined adjacency matrixes by various GSR algorithms on Cora and Citeseer datasets.}

\label{fig8}
\end{figure}

To enhance the comprehension of refined topology, we present the visualization results of edge weights for both the raw and refined graph, as depicted in Fig.~\ref{fig8}.
We adhere to the previous strategy \cite{li2023gslb} to select several subgraphs from Cora and Citeseer datasets. Randomly sampling 20 labeled (L) and 20 unlabeled (U) nodes, we extract four subgraphs and separate them with red lines. 
A subgraph contains two classes, each with 10 nodes. 
Intra- and inter-class connections are separated by green lines. The diagonal elements represent self-loops. 
Comparing the sparse intra- and inter-class connections of raw graph, the refined graph shows a significantly denser structure. However, a denser graph does not necessarily equate to improved performance. 
We find that ECL-GSR maintains a lower frequency of inter-class connections than intra-class ones. This observation aligns with the basic principle of ECL, which is to pull close similar semantic information and push away dissimilar ones.

\section{Conclusion}
In this paper, we advance an Energy-based Contrastive Learning approach to guide GSR and introduce a novel ECL-GSR framework, which jointly optimizes graph structure and representation. ECL is capable of approximating the joint distribution to learn good representations by combining generative and discriminative paradigms. We evaluate the proposed method on the graph node classification task. Experimental results verify its superior effectiveness, efficiency, and robustness.

\section{Acknowledgments}
The work was supported by the National Key Research and Development Program of China (Grant No. 2023YFC3306401).
This work was also supported by the National Natural Science Foundation of China (Grant No. U20B2042 and 62076019).
\bibliography{aaai25}

\clearpage
\section{Appendix}
In the appendix, we discuss critical elements excluded from the main text due to space limitations. We provide a missing proof clarifying the ECL objective, examine significant technologies in related work, introduce the datasets briefly, detail the baseline methods, present further statistical analysis, apply ECL-GSR in graph classification task, and demonstrate the SGLD's iterations and training stability of ECL.

 \subsection{Missing Proof}
In validating both the generative term and discriminative term of ECL objective, we begin by establishing that $p_\theta(\nu)$ represents the marginal distribution of $\nu$ and demonstrating that it is an EBM:
\begin{align*}
    \frac{\exp(-E_\theta(\nu))}{Z(\theta)}\ &= \frac{1}{Z(\theta)}\int{e}^{-\Vert z-z^\prime\Vert^2/\tau} d\nu^\prime \\
		&= \int\frac{1}{Z(\theta)}{e}^{-\Vert z-z^\prime\Vert^2/\tau} d\nu^\prime \\
		&= \int p_\theta(\nu,\nu^\prime) d\nu^\prime \\
		&= p_\theta(\nu).
\end{align*}
At the same time, leveraging the following relation: $\log p_\theta(\nu)=-\log Z(\theta)-E_\theta(\nu)$, we establish the Eq. \ref{eq11}:
\begin{align*}
	\nabla_{\theta}\log p_\theta(\nu) &= -\frac{1}{Z(\theta)}\nabla_{\theta}Z(\theta) - \nabla_{\theta}E_\theta(\nu) \\
    &= -\frac{1}{Z(\theta)}\nabla_{\theta} \int{e}^{-E_{\theta}(\nu)}d\nu-\nabla_{\theta}E_{\theta}(\nu) \\
    &= -\frac{1}{Z(\theta)} \int\nabla_{\theta}{e}^{-E_{\theta}(\nu)}d\nu-\nabla_{\theta}E_{\theta}(\nu) \\
    &= -\frac{1}{Z(\theta)} \int[-\nabla_{\theta}E_\theta(\nu)]\cdot{e}^{-E_{\theta}(\nu)}d\nu-\nabla_{\theta}E_{\theta}(\nu) \\
    &= \int[\nabla_{\theta}E_\theta(\nu)]\cdot\frac{1}{Z(\theta)}{e}^{-E_{\theta}(\nu)}d\nu-\nabla_{\theta}E_{\theta}(\nu) \\
	&= \int[\nabla_{\theta}E_\theta(\nu)]\cdot p_\theta(\nu) d\nu -\nabla_{\theta}E_{\theta}(\nu)\\
    &= \mathbb{E}_{p_\theta}[\nabla_\theta E_\theta(\nu)] - \nabla_{\theta} E_{\theta}(\nu).
\end{align*}
Finally, by employing Bayes’ rule, we establish the gradient of the objective of ECL in Eq. \ref{eq13}.

\subsection{Related Work}
\paragraph{\textbf{Energy-based Models}} EBMs \cite{lecun2006tutorial} are generative methods aiming to learn an energy function $E_\theta(\chi)$ that assigns low energy values to inputs $\chi$ by directly maximizing the log-likelihood of the joint distribution \cite{grathwohl2019your}.
EBMs have effectively resolved various downstream challenges across multiple domains, including visual representation learning \cite{qian2021spatiotemporal} and the generation of images or text \cite{du2019implicit, qin2022cold}. \cite{yang2023m} introduce a Manifold EBM, which enhance the training process by using a simplified, informative initialization that closely aligns with the data manifold. This approach significantly reduces the need for extensive sampling, thereby speeding up training and improving stability while maintaining high generative performance. \cite{roy2023gad} utilize the Jarzynski Equality, which enables efficient estimation of the cross-entropy gradient without the biases typically introduced by the contrastive divergence algorithm used in traditional EBMs' training. Additionally, \cite{roy2023gad} develop a GAD-EBM method that employs EBMs to estimate graph distributions. It efficiently computes likelihoods for rooted subgraphs and uses these likelihoods to accurately detect anomalies within graph nodes. 
\paragraph{\textbf{Contrastive Learning}} CL as a discriminative method trains neural networks by maximizing the agreement between different augmentations of the same data, such as SimCLR \cite{chen2020simple}, MoCo \cite{he2020momentum}, etc. CL has proven effective across diverse applications, including large language models \cite{gunel2020supervised}, video question answering \cite{zhang2022erm}, and action recognition \cite{singh2021semi}. 
\cite{wang2022rethinking} propose an innovative approach to enhance contrastive learning by optimizing the information content of representations. It demonstrates that fine-tuning the strategies for information retention can significantly boost the generalization capabilities and robustness of models, as supported by both theoretical and empirical evidence.
\cite{xu2022contrastive} introduce a contrastive novelty-augmented learning framework, which employs large language models to generate and train on out-of-distribution examples, reducing overconfidence in text classifiers on novel classes without affecting accuracy on known data.
Furthermore, \cite{bo2024graph} present a spectral-spatial contrastive framework that integrates traditional spatial graph views with spectral features. It offers a more stable and scalable process for representation learning.

\paragraph{\textbf{Graph Structure Refinement}} GSR has emerged as a critical research area, especially in developing machine learning models that effectively infer and optimize graph structures from data. Most GSR methods are based on discriminative architectures.
In recent years, \cite{ta2022adaptive} treat the entire graph as learnable parameters and optimize the parameters with an adaptive structure learning component from the macro and micro perspectives for traffic forecasting. \cite{pan2022aagcn} define a metric learning approach including Mahalanobis distance and non-linear mapping to measure node relationships in the graph for person re-identification. \cite{zhao2023self} estimate the underlying probabilities of edges by a multi-view contrastive learning framework and refine links via a sampling process from certain distributions comprehensively.
However, these approaches always necessitate complex pipeline designs tailored for particular applications, which greatly limits the flexibility and applicability of the frameworks.
To date, a combination of generative and discriminative method has not yet appeared in graph representation learning and GSR.

%In the context of GSL, both generative and discriminative methods play crucial roles. Generative methods, such as those described by Kipf and Welling (2016), focus on modeling the joint probability distribution of the graph's nodes and edges, facilitating the generation of new graph samples. These methods are often realized through variational autoencoders (VAEs) which are adept at capturing the latent variables that represent underlying graph structures. On the other hand, discriminative methods, exemplified by Hamilton et al. (2017), aim to predict labels for graph nodes or entire graphs by learning from the features and structure of the graph directly, rather than modeling the entire data distribution. Such methods typically leverage Graph Neural Networks (GNNs) to learn powerful node embeddings that capture both local structure and node feature information. The interplay between these approaches enhances our ability to perform tasks such as link prediction, node classification, and graph classification, pushing the boundaries of what is achievable with GSL.

\begin{table*}[ht]
\begin{center}  
\resizebox{\linewidth}{!}{ 
\begin{tabular}{ccccccccc}
\toprule  %添加表格头部粗线 
\multirow{2}{*}{\textbf{Method}}    & \multicolumn{2}{c}{\textbf{Cora}} & \multicolumn{2}{c}{\textbf{Citeseer}} & \multicolumn{2}{c}{\textbf{Actor}} & \multicolumn{2}{c}{\textbf{Pubmed}} \\
& \textit{T-statistic}  & \textit{P-value} & \textit{T-statistic}  & \textit{P-value} & \textit{T-statistic}  & \textit{P-value} & \textit{T-statistic}  & \textit{P-value}      \\
\midrule
GCN       & 12.33  & $4.24\times10^{-9}$  & 7.85   & $4.18\times10^{-6}$  & 11.45  & $1.40\times10^{-8}$  & 5.46    & $3.66\times10^{-3}$ \\
GAT       & 11.15  & $2.12\times10^{-8}$  & 11.00  & $2.62\times10^{-8}$  & 14.54  & $2.83\times10^{-10}$ & 6.89    & $2.51\times10^{-5}$ \\
LDS       & 6.41   & $6.37\times10^{-5}$  & 11.62  & $8.97\times10^{-7}$  & 10.14  & $3.96\times10^{-3}$  & --      & -- \\
GEN       & 5.66   & $2.97\times10^{-4}$  & 8.22   & $4.65\times10^{-4}$  & 14.72  & $2.31\times10^{-10}$ & 8.51    & $3.22\times10^{-4}$ \\
SGSR      & 10.21  & $6.84\times10^{-4}$  & 5.24   & $1.55\times10^{-2}$  & 13.35  & $1.15\times10^{-9}$  & 4.71    & $2.29\times10^{-3}$ \\
GRCN      & 6.95   & $1.12\times10^{-3}$  & 13.06  & $2.04\times10^{-5}$  & 11.71  & $9.69\times10^{-9}$  & 4.21    & $6.86\times10^{-3}$ \\
IDGL      & 7.58   & $2.47\times10^{-3}$  & 3.99   & $8.05\times10^{-2}$  & 13.40  & $1.08\times10^{-9}$  & --      & -- \\
GAuG-O    & 5.92   & $1.71\times10^{-4}$  & 4.22   & $6.67\times10^{-3}$  & 16.83  & $2.42\times10^{-11}$ & 9.10    & $8.02\times10^{-3}$ \\
SUBLIME   & 8.09   & $8.85\times10^{-4}$  & 7.49   & $4.86\times10^{-4}$  & 8.33   & $1.77\times10^{-6}$  & 16.04   & $5.45\times10^{-11}$ \\
ProGNN    & 16.33  & $3.99\times10^{-11}$ & 18.10  & $6.97\times10^{-12}$ & 25.95  & $1.34\times10^{-14}$ & 21.63   & $3.22\times10^{-13}$ \\
CoGSL     & 13.06  & $1.66\times10^{-9}$  & 4.67   & $3.67\times10^{-2}$  & 12.37  & $3.81\times10^{-5}$  & --      & -- \\
STABLE    & 15.52  & $9.45\times10^{-11}$ & 9.43   & $2.86\times10^{-7}$  & 7.58   & $6.86\times10^{-6}$  & --      & -- \\
NodeFormer& 12.26  & $4.62\times10^{-9}$  & 6.09   & $1.22\times10^{-4}$  & 14.36  & $3.46\times10^{-10}$ & 4.90    & $1.49\times10^{-3}$ \\
\bottomrule 		
\end{tabular}
}
\caption{\label{table3}Pairwise independent sample $t$-tests comparison of ECL-GSR with other methods on Cora, Citeseer, Actor, and Pubmed datasets.} 
\end{center}
\end{table*}

\subsection{Statistical Analysis}
ECL-GSR achieves the state-of-the-art performance across eight datasets using a uniform set of hyperparameters, demonstrating its effectiveness, generalization, and robustness. To bolster the credibility of our experimental results, we conduct a pairwise independent sample $t$-tests comparison of ECL-GSR with other methods on four datasets. The obtained P-values, all below 0.05, signify a statistically significant difference between our framework and the compared methods. Detailed findings are presented in Table \ref{table3}.

\begin{table}[h]
\begin{center} 
\resizebox{\linewidth}{!}{ 
\begin{tabular}{ccccccccccc}
\toprule  %添加表格头部粗线 
\textbf{Model} & \textbf{COLLAB} & \textbf{MUTAG} & \textbf{PROTEINS} \\ 
\midrule[0.8pt]
GCN            & 76.96\scriptsize{$\pm$2.28}        & 73.92\scriptsize{$\pm$8.84}   & 67.52\scriptsize{$\pm$6.71}    \\      
GAT            & 79.08\scriptsize{$\pm$1.36}        & 78.71\scriptsize{$\pm$7.51}   & 68.63\scriptsize{$\pm$6.24}    \\      
SAGE           & 75.58\scriptsize{$\pm$2.04}        & 68.65\scriptsize{$\pm$4.31}   & 64.47\scriptsize{$\pm$7.15}    \\     
VIB-GSL        & 77.14\scriptsize{$\pm$1.59}        & 68.63\scriptsize{$\pm$5.15}   & 65.68\scriptsize{$\pm$8.53}    \\      
HGP-SL         & 78.06\scriptsize{$\pm$2.17}        & 78.07\scriptsize{$\pm$10.85}  & 70.80\scriptsize{$\pm$4.25}    \\      

PSCN           & 72.60\scriptsize{$\pm$2.15}        & 88.95\scriptsize{$\pm$4.37}   & 75.00\scriptsize{$\pm$2.51}    \\ 
DGCNN          & 73.76\scriptsize{$\pm$0.49}        & 85.83\scriptsize{$\pm$1.16}   & 75.54\scriptsize{$\pm$0.94}    \\      
CapsGNN        & 79.62\scriptsize{$\pm$0.91}        & 86.67\scriptsize{$\pm$6.88}   & 76.28\scriptsize{$\pm$3.63}    \\      
GFN-light      & 81.34\scriptsize{$\pm$1.73}        & 89.89\scriptsize{$\pm$7.14}   & 77.44\scriptsize{$\pm$3.77}    \\      
\midrule[0.5pt]
ECL-GSR    & \textbf{{83.00\scriptsize{$\pm$1.43}}} & \textbf{{91.45\scriptsize{$\pm$2.75}}} & \textbf{{78.77\scriptsize{$\pm$2.36}}} \\ 
\bottomrule 		
\end{tabular}
}
\caption{\label{table4}Graph classification accuracy (mean(\%)±std) on various benchmark datasets. \textbf{Bold} indicates the best performance.}  
\end{center}
\end{table}

\subsection{Graph Classification}
%We deduce a tractable approximation for the ECL objective, which facilitates the training stability. This subsection presents an analysis of the convergence properties of ECL-GSR applied to the Cora and Citeseer datasets, employing a learning rate of 0.001. The total ECL objective function is denoted as $\mathcal{L}_{E}(\theta)=\mathcal{L}_b(\theta)+\beta\mathcal{L}_r(\theta)$, where $\mathcal{L}_b(\theta)$ encompasses both the discriminative term $\mathbb{E}_{p_d}[-\log{p_\theta(\nu|\nu^\prime)}]$ and the generative term $\alpha{\mathbb{E}}_{p_d}[-\log{p_\theta(\nu)}]$, while $\beta\mathcal{L}_r(\theta)$ represents the $L_2$ regularization loss. The hyperparameters $\alpha$ and $\beta$ are set to 0.1 and 0.001, respectively. The dashed lines signify the mean value of the final 10 epochs, marking the convergence point of ECL-GSR. As illustrated in Fig.~\ref{fig8}, ECL-GSR converges steadily, showing the effectiveness of the approximate implementation.ECL-GSR converges steadily, showing the effectiveness of the approximate implementation.ECL-GSR converges steadily, showing the effectiveness of the approximate implementation.ECL-GSR

To evaluate the efficacy of ECL across diverse downstream tasks, we further expand our analysis to include three distinct graph classification datasets. 
COLLAB \cite{yanardag2015deep} is employed to examine networks of scientific research collaboration, MUTAG \cite{debnath1991structure} features graphical depictions of chemical molecules, primarily for assessing the mutagenicity of chemical compounds, and PROTEINS \cite{borgwardt2005protein} utilizes graphical representations of protein structures to facilitate the prediction of protein functions.
We test a total of nine methods. GCN, GAT, and SAGE served as the baseline methods. VIB-GSL \cite{sun2022graph} and HGP-SL \cite{zhang2019hierarchical} are approaches predicated on GSR, while PSCN \cite{niepert2016learning}, DGCNN \cite{zhang2018end}, CapsGNN \cite{xinyi2018capsule}, and GFN-light \cite{chen2019powerful} represent graph classification methods developed using an end-to-end architecture. As illustrated in Table \ref{table4}, ECL-GSR outperforms the others in terms of accuracy, thereby demonstrating that the flexibility of ECL on graph-related downstream tasks.

\subsection{SGLD's Iterations}
\begin{figure}[h]
\centering
\includegraphics[width=\linewidth]{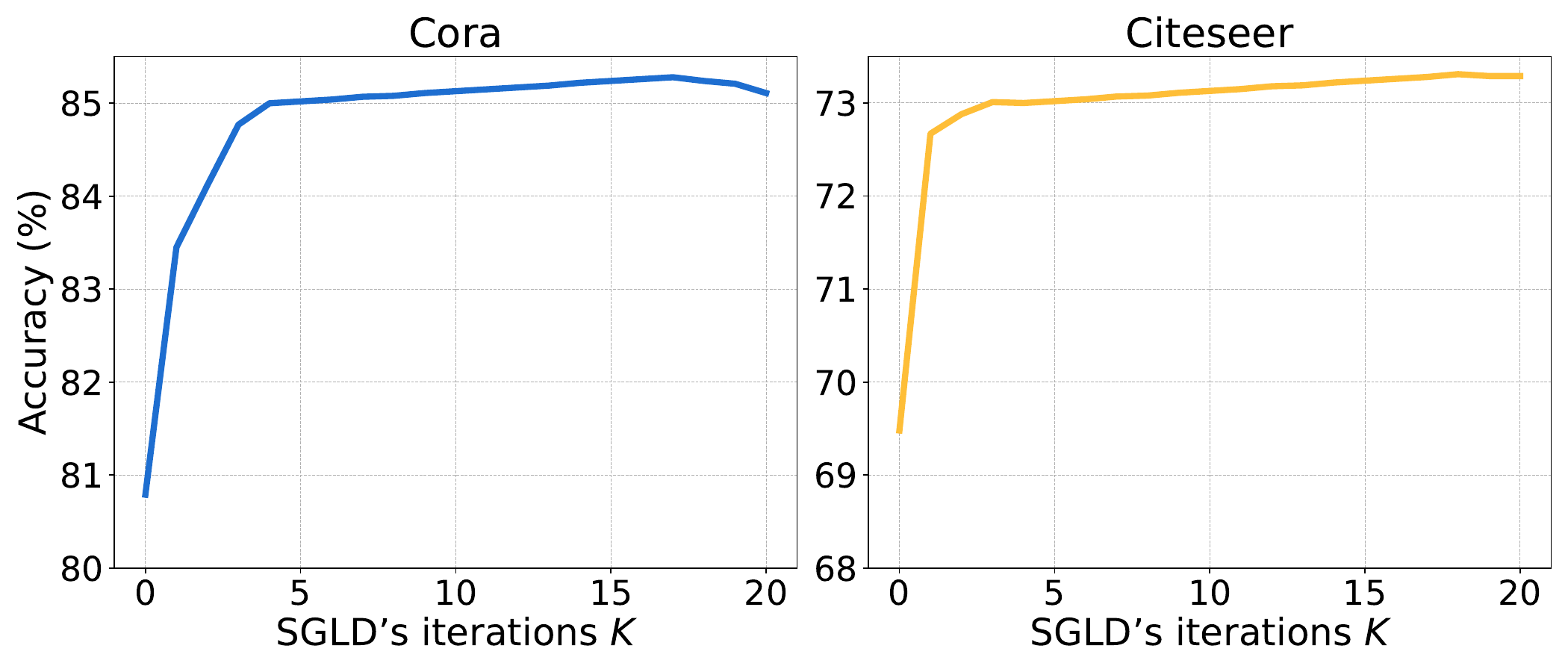} 
\caption{The number of SGLD iterations analysis on Cora and Citeseer datasets.}
\label{fig9}
\end{figure}

We examine the effects of varying the number of SGLD iterations on the performance of ECL-GSR, as depicted in Fig.~\ref{fig9}. It is observed that a few steps of SGLD iterations swiftly elevates the node classification accuracy to a substantial level. However, further increases in the number of iterations yield only marginal improvements in accuracy, demonstrating a significant marginal effect. With the consequent increase in computational time, prolonged iterations is unacceptable. Therefore, we opt for $K=3$ iterations to trade-off between performance and computational speed.

\subsection{Training Stability}
\begin{figure}[h]
\centering
\includegraphics[width=\linewidth]{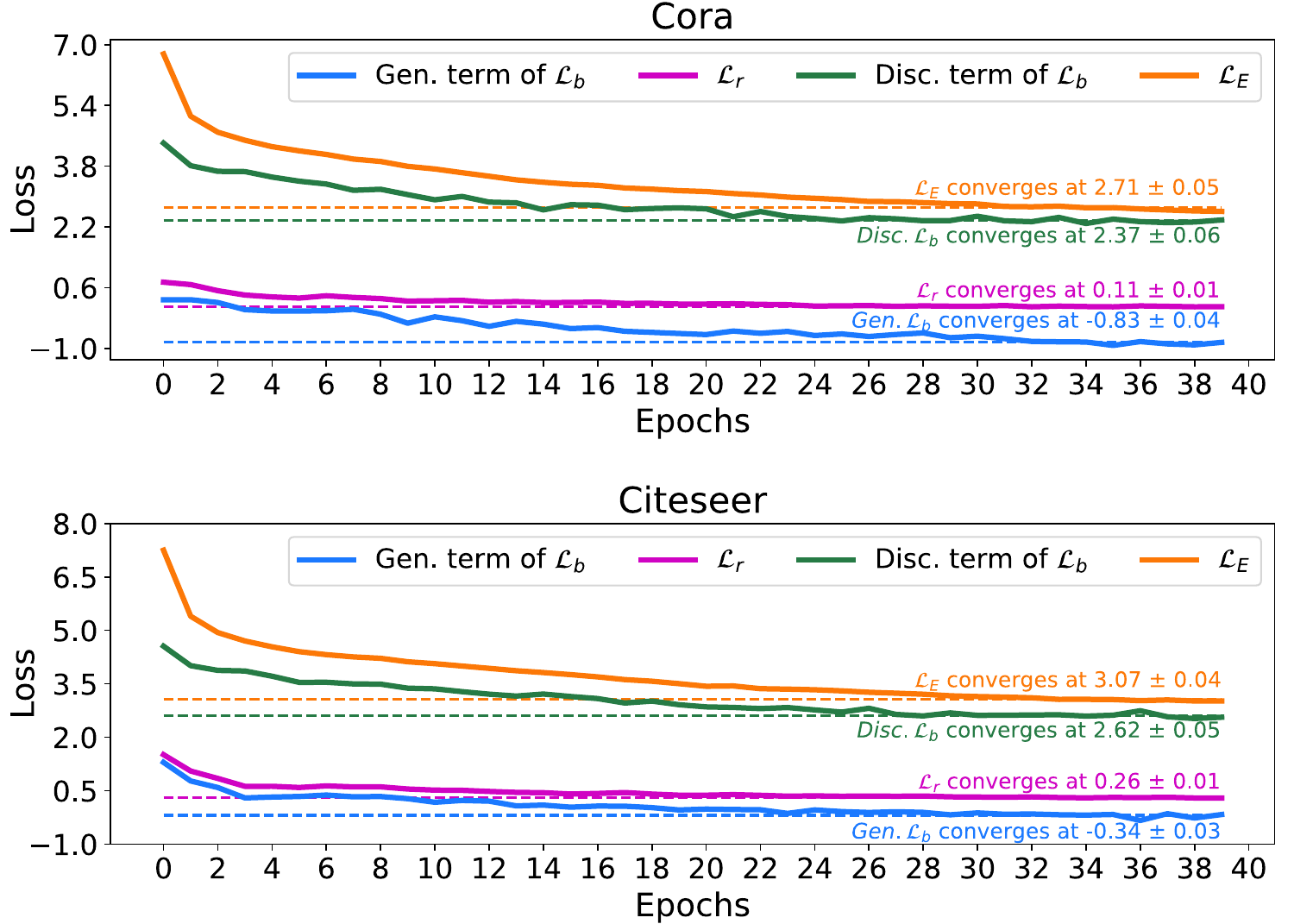} 
\caption{Training dynamics of ECL-GSR with multiple losses on Cora and Citeseer datasets.}
\label{fig10}
\end{figure}

We deduce a tractable approximation for the ECL objective, which facilitates the training stability. This subsection presents an analysis of the convergence properties of ECL-GSR applied to the Cora and Citeseer datasets, employing a learning rate of 0.001. The total ECL objective function is denoted as $\mathcal{L}_{E}(\theta)=\mathcal{L}_b(\theta)+\beta\mathcal{L}_r(\theta)$, where $\mathcal{L}_b(\theta)$ encompasses both the discriminative term $\mathbb{E}_{p_d}[-\log{p_\theta(\nu|\nu^\prime)}]$ and the generative term $\alpha{\mathbb{E}}_{p_d}[-\log{p_\theta(\nu)}]$, while $\beta\mathcal{L}_r(\theta)$ represents the $L_2$ regularization loss. The hyperparameters $\alpha$ and $\beta$ are set to 0.1 and 0.001, respectively. The dashed lines signify the mean value of the final 10 epochs, marking the convergence point of ECL-GSR. As illustrated in Fig.~\ref{fig10}, ECL-GSR converges steadily, showing the effectiveness of the approximate implementation.

\begin{table*}[ht]
\begin{center} 
\begin{tabular}{lccccccccc}
\toprule  %添加表格头部粗线 
Datasets & Cora & Citeseer  & Cornell & Texas & Wisconsin & Actor & Pubmed & OGB-Arxiv\\
\midrule[0.8pt]
\#Nodes    & 2708  & 3,327   & 183   & 183  & 251   & 7600   & 19717   & 169343   \\
\#Edges    & 5278  & 4614    & 277   & 279  & 450   & 26659  & 44325   & 1157799  \\
\#Classes  & 7     & 6       & 5     & 5    & 5     & 5      & 3       & 40       \\
\#Features & 1,433 & 3,703   & 1,703 & 1703 & 1,703 & 932    & 500     & 767      \\
\#Homophily & 0.81 & 0.74    & 0.12  & 0.06 & 0.18  & 0.22   & 0.80    & 0.65     \\
\#Degree   & 3.90  & 2.77    & 3.03  & 3.05 & 3.59  & 7.02   & 4.50    & 13.67    \\
\#Train    & 140   & 120     & 87    & 87   & 120   & 3648   & 60      & 90941    \\
\#Validation & 500 & 500     & 59    & 59   & 80    & 2432   & 500     & 29799    \\
\#Test     & 1000  & 1000    & 37    & 37   & 51    & 1520   & 1000    & 48603    \\
\bottomrule 		
\end{tabular}
\caption{\label{table5} The statistics of node classification datasets.}
\end{center}
\end{table*}
\subsection{Datasets}
The datasets in our benchmark, all previously released for graph learning or new graph-based tasks, span a range of applications: citation and social networks, website, computer vision, and co-occurrence networks. Detailed statistics for these varied datasets are presented in Table \ref{table5}. We select these benchmark datasets for their broad use in GNN model development and evaluation, as well as their diverse graph representations, which range from small to large. The following sections describe each dataset in detail:

  \begin{itemize}
  	\item \textbf{Cora, Citeseer, Pubmed} The Cora, Citeseer \cite{sen2008collective}, and Pubmed \cite{namata2012query} datasets construct bibliometric networks, with nodes depicting individual scientific papers and connecting edges representing citations. publications are detailed by binary word vectors that flag the non-occurrence or occurrence of specific words from a predefined dictionary. Each node is linked to a distinct one-hot label which serves in the classification task to ascertain the category of the related publication.
  	\item \textbf{Cornell, Texas, Wisconsin} The Cornell, Texas, and Wisconsin collections \cite{pei2020geom} are sourced from WebKB3 by Carnegie Mellon University, encompass web pages from university computer science departments, with nodes and edges representing web pages and hyperlinks, respectively. These are characterized using bag-of-words from the pages and categorized into five distinct classes: student, project, course, staff, and faculty.
  	\item \textbf{Actor} The Actor dataset \cite{tang2009social} is a specialized subgraph focused on actors, extracted from a larger film industry network. It captures the interconnectedness of actors based on their co-occurrence on Wikipedia pages. The features of each actor are distilled into a bag-of-words vector derived from the textual content of their Wikipedia articles.
  	\item \textbf{OGB-Arxiv} The OGB-Arxiv dataset \cite{hu2020open} is part of the Open Graph Benchmark collection, a vast citation network dataset tailored for the rigorous evaluation of graph neural network models. It encompasses a comprehensive set of arXiv papers categorized under the computer science domain. Each paper is represented by a node, with directed edges indicating citation links. The papers are characterized by 128-dimensional word embeddings, which are computed from the titles and abstracts, providing a rich representation of the paper's content.
	\end{itemize}

\subsection{Baselines}
To provide a comprehensive and detailed comparison, we integrate a series of pioneering frameworks and methods that advance the robustness, efficiency, and accuracy across various applications. The following descriptions offer insights into each approach:
\begin{itemize}
\item \textbf{LDS} \cite{franceschi2019learning} LDS is an early work in GSR. It proposes to solve a bilevel program to learn the graph structure and parameters of GNNs jointly. LDS does not assume a fixed adjacency matrix; instead, it models each potential edge as a Bernoulli random variable, making the process probabilistic and allowing for a more flexible and potentially more accurate representation of the underlying data structure.
\item \textbf{GRCN} \cite{yu2021graph} GRCN introduces a graph revision module that performs edge addition and reweighting, allowing the model to predict unseen edges and adaptively revise the graph according to downstream tasks. This approach is efficient in scenarios where graphs are highly incomplete or the available training labels are sparse. The GRCN framework significantly enhances prediction accuracy for downstream tasks by overcoming issues related to over-parameterization and the inability to deal with missing edges, which are common drawbacks in previous graph revision methods.
\item \textbf{ProGNN} \cite{jin2020graph} ProGNN addresses the vulnerability of GNNs to adversarial attacks, which can significantly impair their performance. ProGNN defends against such attacks by learning a clean graph structure that adheres to the mentioned properties and simultaneously updating the GNN parameters. The framework has demonstrated the ability to defend against various adversarial attacks. It outperforms other state-of-the-art defense methods, even with heavily perturbed graphs.
\item \textbf{IDGL} \cite{chen2020iterative} IDGL is a comprehensive framework designed for graph neural networks, which focuses on the simultaneous and iterative learning of both graph structure and graph embedding. The main principle behind IDGL is to enhance the quality of the graph structure through improved node embeddings and, conversely, to improve node embeddings through a better graph structure. This is achieved through an iterative process that refines the graph structure and node embeddings in tandem. The iterative process is designed to terminate when the graph structure is sufficiently optimized for the specific downstream task.
\item \textbf{CoGSL} \cite{liu2022compact} By introducing the contrastive invariance theorem, CoGSL provides insights into the general rules of graph augmentation and elucidates the mechanisms by which GCL achieves its effectiveness. The theoretical foundation of CoGSL is built upon optimizing primary and final views based on mutual information while maintaining performance on labels. The final view obtained through this process is a minimal sufficient structure for the graph. This reflects GSR's growing interest and importance in machine learning and data science, as it provides a means for optimizing graph structure and learning appropriate embeddings.
\item \textbf{SUBLIME} \cite{liu2022towards} SUBLIME utilizes self-supervised contrastive learning to optimize the topology of a graph. Specifically, it generates a target from the original data, called the "anchor graph." It employs a contrastive loss function to maximize the similarity, or agreement, between the anchor graph and the generated graph topology. The critical innovation of SUBLIME lies in its ability to optimize the learned graph topology using the data itself without relying on any external guidance, such as labels, making it a more practical paradigm for GSR. This approach allows for a more data-driven process to refine graph structures, particularly valuable in unsupervised learning scenarios where labeled data may not be readily available.
\item \textbf{STABLE} \cite{li2022reliable} STABLE is unsupervised, meaning it does not rely on pre-labeled training data. The framework's primary advantage lies in its ability to significantly enhance the generalization capacity of graph structures, which is crucial for producing robust models that perform well on unseen data. By focusing on learning stable and fair representations, particularly within the output of GNNs, the STABLE method ensures that the learned representations are fair and stable across different environments and datasets.
\item \textbf{NodeFormer} \cite{wu2022nodeformer} NodeFormer leverages an innovative all-pair message passing mechanism that efficiently propagates node signals across all nodes within a graph. This system is scalable and can handle graphs with node counts ranging from thousands to millions, which is a significant advancement in handling large-scale data in graph neural networks. The model's efficiency is partly due to a specialized latent structure learning approach, which enables NodeFormer to scale its operations to accommodate the expansive node levels typically found in large graphs. By implementing a kernelized Gumbel operator, NodeFormer can reduce the computational complexity and facilitate more effective learning of graph structures.
\item \textbf{SGSR} \cite{zhao2023self} SGSR first estimates a pre-train-finetune pipeline where the pre-training phase uses a multi-view contrastive learning framework for estimating the graph structure through link prediction tasks. After refining the graph by adjusting edges based on probabilities from the pre-trained model, the fine-tuning phase optimizes the GNN for specific tasks without further adjusting the graph structure, thus improving scalability and efficiency. This method has demonstrated superior performance on benchmark datasets and has shown to be significantly faster and less resource-intensive than previous GSR approaches.
\item \textbf{GAuG-O} \cite{zhao2021data} GAuG-O employs bilevel optimization to improve the generalization performance of models on datasets. This technique operates by optimizing an augmentation network based on the performance of a downstream classification network, using a validation set for evaluation. The framework particularly shines in its ability to enhance graph classification results without requiring prior domain knowledge, and it has been shown to outperform other methods, such as FLAG augmentation, on specific benchmarks.
\item \textbf{GEN} \cite{wang2021graph} GEN captures graphs' genesis and community structures, mimicking the complex, clustered interactions within real-world networks. Beyond first-order connections, the observation model incorporates multi-order neighborhood information, considering the relationships between a node and its extended network of neighbors, thereby capturing a more intricate web of interactions.
Employing Bayesian inference, GEN updates the probability of the graph's structural hypothesis as it assimilates this expansive neighborhood data. This approach allows for a more nuanced and dynamic inference of graph structure, enhancing the accuracy of predictions related to the network's topology and evolution.
\end{itemize}

\end{document}